\documentclass[10pt,twocolumn,letterpaper]{article}

\usepackage{cvpr}              
\definecolor{cvprblue}{rgb}{0.21,0.49,0.74}
\usepackage[pagebackref,breaklinks,colorlinks,allcolors=cvprblue]{hyperref}
\usepackage{multirow} 
\usepackage{algorithm}
\usepackage{algorithmicx}
\usepackage{algpseudocode}
\usepackage{makecell}
\usepackage[table]{xcolor}
  
\def\paperID{15192} 
\def\confName{CVPR}
\def\confYear{2026}

\newcommand{\method}{CRISP}

\title{Decompose, Mix, Adapt: A Unified Framework for Parameter-Efficient Neural Network Recombination and Compression}

\author{Nazia Tasnim\\
Boston University\\
{\tt\small nimzia@bu.edu}
\and
Shrimai Prabhumoye\\
NVIDIA, Boston University\\
{\tt\small sprabhumoye@nvidia.com}
\and
Bryan A. Plummer\\
Boston University\\
{\tt\small bplum@bu.edu}
}
\begin{document}
\maketitle
\begin{abstract}
Parameter Recombination (PR) methods aim to efficiently compose the weights of a neural network for applications like Parameter-Efficient FineTuning (PEFT) and Model Compression (MC), among others. Most methods typically focus on one application of PR, which can make composing them challenging.  For example, when deploying a large model you may wish to compress the  model and also quickly adapt to new settings.  However, PEFT methods often can still contain millions of parameters. This may be small compared to the original model size, but can be problematic in resource constrained deployments like edge devices, where they take a larger portion of the compressed model's parameters.  To address this, we present Coefficient-gated weight Recombination by Interpolated Shared basis Projections (\method{}), a general approach that seamlessly integrates multiple PR tasks within the same framework.  \method{} accomplishes this by factorizing pretrained weights into basis matrices and their component mixing projections. Sharing basis matrices across layers and adjusting its size enables us to perform MC, whereas the mixer weight's small size (fewer than 200 in some experiments) enables \method{} to support PEFT.  Experiments show \method{} outperforms methods from prior work capable of dual-task applications by 4-5\% while also outperforming the state-of-the-art in PEFT by 1.5\% and PEFT+MC combinations by 1\%.
Our code is available on the repository: \url{https://github.com/appledora/CRISP-CVPR26}.

\end{abstract}   
\section{Introduction}
\label{sec: intro}
\begin{figure}[t]
    \centering
    \includegraphics[width=\linewidth]{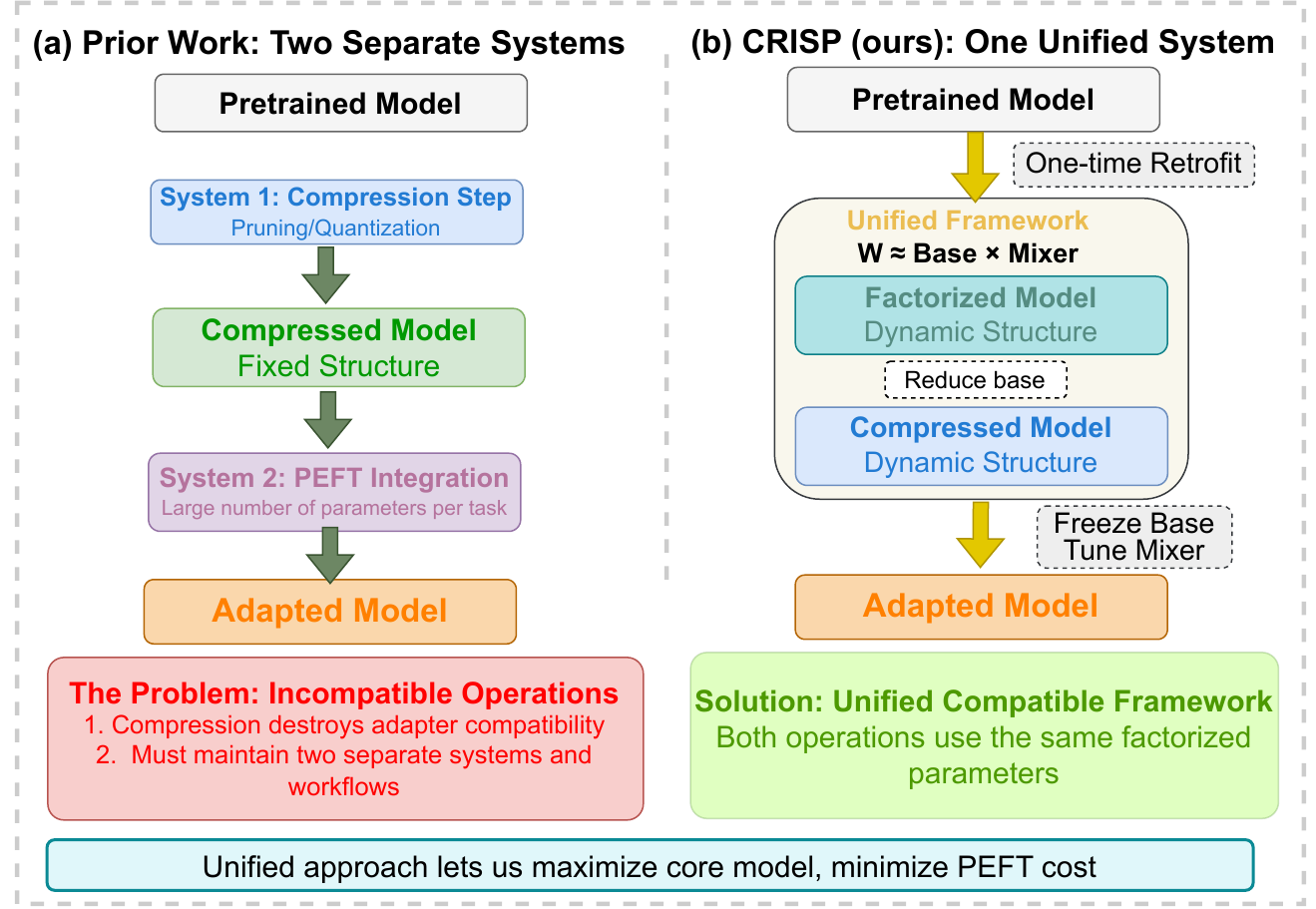}
    \vspace{-4mm}
    \caption{\textbf{PR approach comparison.} \textbf{(a)} Prior work in PR typically focuses on PEFT or MC alone~\cite{Wang2024NeuralNP, Zhang_Luo_Yu_Li_Lin_Ye_Zhang_2024, Erko2023HyperDiffusionGI,hu2022lora,10.5555/3692070.3693369,si2024see,10.5555/3600270.3601482, liang2024inflora, 10635615, pmlr-v235-nikdan24a,Ahmed_2025_CVPR, MiniViT, hao2022manifold, Rangwani_2024_CVPR,10678046,wang2025basis,glandorf2025p3b, pmlr-v202-shi23e, 10.1007/s11432-022-3646-6}, which can result in efficient combinations when deployed together. \textbf{(b)} Our unified PR approach \method{} decomposes a pretrained models weights that support both MC and PEFT, enabling us to more effectively use parameter budgets even as tasks scale.{\vspace{-1em}}}
    \label{fig:method-comparison}
\end{figure}

The sheer scale of large transformer models has triggered a wave of tasks that aim to support their deployment in more computationally constrained environments, such as parameter-efficient finetuning (PEFT)~\cite{Wang2024NeuralNP, Zhang_Luo_Yu_Li_Lin_Ye_Zhang_2024, Erko2023HyperDiffusionGI,hu2022lora,10.5555/3692070.3693369,si2024see,10.5555/3600270.3601482, liang2024inflora, 10635615, pmlr-v235-nikdan24a}, Model Compression (MC)~\cite{Ahmed_2025_CVPR, MiniViT, hao2022manifold, Rangwani_2024_CVPR,10678046,wang2025basis,glandorf2025p3b, pmlr-v202-shi23e, 10.1007/s11432-022-3646-6}, and token pruning~\citep{li2023llama,yao2024deco,chen2024image,zhang2024sparsevlm,10855481}, among others.  As shown in \cref{fig:method-comparison}(a), methods for tasks like PEFT and MC often use some type of Parameter Recombination (PR), where the parameters of the model are reconfigured to either enable low-resource adaptation to one or more tasks, or reduce the model size, respectively. However, many applications, especially those on edge devices like robotics or mobile phones, require models with small storage requirements and need to be easily adaptable to new settings.  Despite this, these two tasks are typically explored in isolation, which can result in inefficient combinations when put together.  For example, if we use a MC method to reduce the parameters of a ViT-S/16 by 50\%, then applying PEFT method DoRA~\cite{10.5555/3692070.3693369} on the VTAB-1K dataset~\cite{zhai2019large} would make the compressed model 19\% larger, removing a significant portion of MC's benefit.

To address these issues, we propose Coefficient-gated weight Recombination by Interpolated Shared basis Projections (\method{}), a novel factorized basis-mixer reparameterization scheme that supports both PEFT and MC in the same PR framework. As illustrated in \cref{fig:method-comparison}(b), \method{} introduces learnable Factorized Basis matrices that generate weight matrices through parametric transformations with mixer matrices, enabling systematic parameter sharing across transformer layers.  Controlling the size of the basis matrices and how many layers share them enables \method{} to reduce the number of parameters required to store the model.  Additionally, finetuning with a frozen basis matrix and only updating the mixer matrices allows \method{} to also support task-dependent adaptation capabilities. As our experiments will show, our approach is still effective when using few trainable parameters per task ($<200$). 

\method{} creates its basis and mixer matrices during a short adaptation phase performed on pretrained weights, which can be completed in $< 1$ minute for ViT models~\cite{DBLP:journals/corr/abs-2010-11929} models or $< 30$ minutes for large Llama models~\cite{li2023llama,dubey2024llama3}. 
We find that first creating an uncompressed decomposed model provides a more effective teacher for any subsequent compression.  This multi-stage compression scheme is reminiscent of weight pruning methods that would reduce a model in stages (\eg,~\cite{Iurada_2024_CVPR,NEURIPS2024_c1c44e46,10.5555/3737916.3742403}), but with a very low computational cost as noted above.

Our approach most closely resembles RECAST~\cite{tasnimRECAST2025}, which can be seen as a special case of the unconstrained version of our method where the mixing weights are restricted to be a vector (whereas in our approach they are represented as a matrix). This limitation means that RECAST only sees benefits when a very small number of parameters were used for finetuning ($<200$).  In contrast, our more general formulation provides benefits over a wide range of parameter budgets while still reporting gains in settings RECAST does well.  As a result, \method{} outperforms RECAST by 4-6\% with negligible differences in computational complexity.  In addition, \method{} reports a 1.5\% gain over the state-of-the-art in PEFT and also improves performance by almost 1\% compared to combinations of PEFT and MC methods.

Our key contributions include: 
\begin{itemize}
    \item  We introduce Coefficient-gated weight Recombination by Interpolated Shared basis Projections (\method{}),  which supports both PEFT and MC in a unified PR framework.
    \item We propose a carefully designed weight generation strategy that provides benefits over a wider range of configurations than closely related work~\cite{tasnimRECAST2025} while also reducing overfitting without introducing new hyperparameters.
    \item We provide extensive evaluation showing \method{} achieves superior performance compared to existing PEFT methods while using fewer trainable parameters and being faster to compute than most prior work. 
\end{itemize}

\section{Related Work}
\label{sec: related}

Two popular uses of Parameter Recombination (PR) methods are Parameter-Efficient FineTuning (PEFT)~\cite{hu2022lora,ren-etal-2024-melora,valipour-etal-2023-dylora,nikdan2024rosa, pmlr-v235-nikdan24a,tasnimRECAST2025,liu2024boft,li2024vblora,10.5555/3600270.3600412,10.5555/3692070.3693369,liao2024in,zhang2023adaptive}, and Model Compression (MC)~\cite{9859786,Zhang_2024_CVPR,wang2025basis,Ahmed_2025_CVPR,11186163,MiniViT,Ye_2024_CVPR,10549904,swaminathan2020sparse}, which both often optimize the weight space of models to minimize computational resources for either adapting to a dataset or reducing model size, receptively.  These methods are often explored in isolation due to varying tasks goals (\eg, PEFT methods actually increase total model size), but many applications require both PEFT and MC to create small models that can be efficiently adapted to new tasks, such as in robotics, mobile phones, or other edge devices.  While we can combine methods from each of these tasks (\eg, by compressing first and then applying a PEFT method for adaptation), we show these compositions underperform our more general PR formulation that supports both PEFT and MC.

Some PR methods are functionally capable of both PEFT and MC, but were often developed and evaluated for only one task and modality.  These include template mixing methods~\cite{savarese2018learning,Plummer2020NeuralPA,tasnimRECAST2025} or those based on SVD~\cite{svdiff2023cvpr,wang2025svdllm,10.3390/s21165599}. For example, the MC method Basis Sharing~\cite{wang2025basis} proposes a mixing procedure inspired by SVD decomposition, that acts functionally similar to template mixing methods.  This provides an efficient MC method for LLMs, yet we found it fails catastrophically on ViT architectures and underperforms when coupled with PEFT methods on LLMs. Further, some methods appear to perform joint compression and model adaptation (\eg,~\cite{zhang-etal-2024-loraprune,chen2023lorashearefficientlargelanguage,grigore2024weight}), but we find these result in specialized methods that do not generalize well.  In contrast, \method{}'s unified framework generalizes across tasks and modalities.

A range of alternative methods related to PEFT or MC have been explored in prior work, including prompt tuning~\cite{wang2022learning,lion2024aaai,10.1007/978-3-031-72995-9_4, 10.5555/3692070.3694125,Park_2024_CVPR,Shang_2025_ICCV,10.1145/3657632,NEURIPS2024_0a0eba34}, parameter pruning~\cite{10470980,10.1007/978-3-031-73404-5_14,10.5555/3737916.3742403,NEURIPS2024_c1c44e46,sengupta2025you,Iurada_2024_CVPR,Agarwal_2024_CVPR,liu-etal-2024-pruning,yang-etal-2024-laco,ZHANG2025107656,khaki2024the}, and knowledge distillation~\cite{hao2022manifold,10678046,Rangwani_2024_CVPR,KANG2024112531,Wang_2024_CVPR,Denize_2024_WACV,multi2024aaai,Liu_2024_CVPR},  among others.  However, they have similar shortcomings as past PR approaches as they evaluate on PEFT or MC alone, and often cannot be applied to both tasks.  For example, prompt tuning cannot be used to compress a model, and knowledge distillation focuses on making a smaller model, but then requires a PEFT method to adapt to new tasks. In contrast, \method{} supports both PEFT and MC, and prior work suggests \method{} can likely be combined to boost performance in their settings~\cite{Plummer2020NeuralPA} (\eg, using both pruning and \method{} together for compression).  We leave the exploration of combining PR and these alternative methods to future work.

\begin{figure*}[t]
    \centering
    \includegraphics[width=\linewidth]{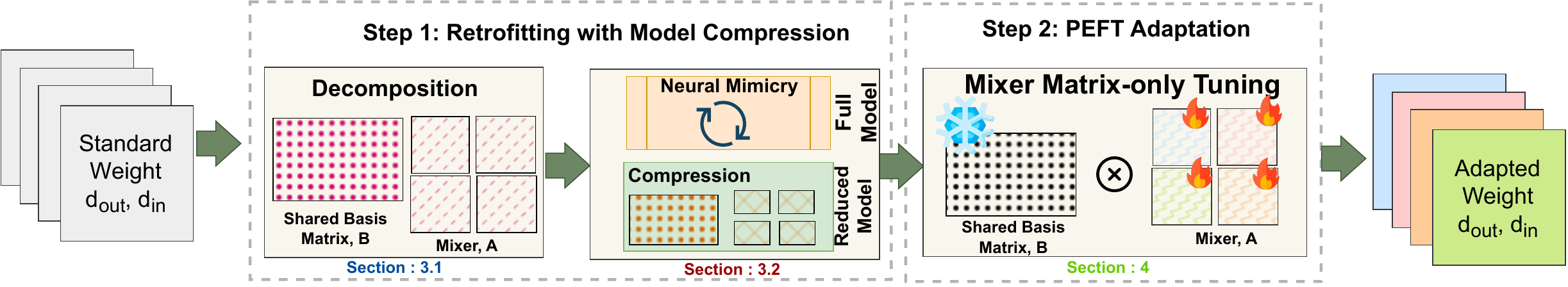}
    \vspace{-4mm}
    \caption{
    \method{} decomposes a pretrained weight matrix into a frozen shared basis 
    and small, learnable mixer matrices, then retrofits these components back 
    into the model (Sec.~\ref{sec:retrofitting}). Compression is achieved by 
    reducing the basis size, while adaptation is 
    enabled by fine-tuning only the lightweight, nonlinearly gated mixer matrices 
    (Sec.~\ref{sec:reparameterization}) - allowing both Parameter Recombination (PR) applications to coexist within a single factorized structure with no redundant adapters).}
    \label{fig:method-flow}
\end{figure*}

\section{\method{} approach}
\label{sec: method}

PR methods aim to use a small number of trainable parameters to produce the weights of a pretrained network to support applications like efficiently adapting a dataset, such as done for PEFT, or to reduce its storage requirements, as done for MC.  More formally, given some trainable parameters $\theta_i$ for layer $i$, PR methods define a transformation $\mathcal{T}$ to produce weights $W_i \in \mathbb{R}^{d_{\text{out}} \times d_{\text{in}}}$, \ie, $W_i=\mathcal{T}(\theta_i)$. There are three major components for PR methods where methods may differ:  1) definition of the transformation $\mathcal{T}$, 2) the role of the pretrained model's weights, and 3) $\theta$ sharing strategy.  
\cref{sec:reparameterization} will define our transformation function $\mathcal{T}_{\method{}}$ and is where our technical novelty lies. \cref{sec:retrofitting} will discuss the sharing strategy and the decomposition-recombination framework we adopt to showcase the benefits of $\mathcal{T}_{\method{}}$, but our approach should be comparable with the choices used by other PR methods as well. See \cref{fig:method-flow} for an overview of our approach

\subsection{Weight Reparameterization via Learnable Basis Modulation}
\label{sec:reparameterization}

Many differences in PR methods are largely due to different definitions of their transformation function.  For example, PEFT method LoRA~\cite{hu2022lora} splits its trainable parameters into a low rank decomposition that is used to adjust the pretrained model's weights, \ie, its transformation for the $i$th layer would be defined as:
\begin{equation}
   \mathcal{T}_{LoRA}(B^r_i,A^r_i) = B^r_iA^r_i + W_{p_i},
   \label{eq:lora}
\end{equation}
where $A^r_i \in \mathbb{R}^{r \times d_{\text{in}}}$,  $W_{p_i}$ and $B^r_i \in \mathbb{R}^{d_{\text{out}} \times r}$ are the trainable parameters $\theta$ and $W_{p_i}$ is the frozen pretrained model's weights.  We use superscripts on matrices $A,B$ to signify the key hyperparmeters to be set by $\mathcal{T}$.  Since \cref{eq:lora} requires pretrained weights and trainable parameters, the total size increases. In contrast, MC method Basis Sharing~\cite{wang2025basis} can reduce the number of parameters by removing the reliance on pretrained weights and adjusting the rank $r$, \ie,
\begin{equation}
   \mathcal{T}_{BasisSharing}(B^r_i,A^r_i) = B^r_iA^r_i. 
\end{equation}

 This represents a small change to $\mathcal{T}$ by itself, but Basis Sharing also changed how the pretrained weights were used.  Instead of adjusting the pretrained weights like LoRA-based methods, Basis Sharing would use the pretrained weights to create the matrices using a scaled SVD decomposition. The scaling factors were estimated using a few dataset samples following a similar procedure used by~\citet{wang2025svdllm}. Another notable difference is that the basis matrix $B^r_i$ would be shared across multiple layers, providing another mechanism to increase the compression rate (by sharing across more layers). Layer-specific weights can still be generated via the layer-specific mixing coefficients $A^r_i$.  While this formulation is sufficient for compression, the matrix $A^r_i$ can still be relatively large, and reducing its size is also tied to additional compression.  Thus, $\mathcal{T}_{BasisSharing}$ is not suitable for PEFT as any improvements by finetuning $A^r_i$ (keeping $B^r_i$ frozen) might be lost due to lower model capacity.  

RECAST~\cite{tasnimRECAST2025} addresses this by defining $a^r$ as a vector of size $r$, ensuring a compact size, and making rows of $B$ the same size as target weights.  In addition,  RECAST averages over a set of $K$ coefficient vectors to enable additional expressivity in the produced weights, \ie,
\begin{equation}
   \mathcal{T}_{RECAST}(B^{*r}_i,a^r_{i,j}) = \frac{1}{K}\sum^K_{j=1} B^{*r}_ia^r_{i,j}, 
   \label{eq:recast}
\end{equation}
where $B^{*r}_i \in \mathbb{R}^{d_{\text{in}}* d_{\text{out}} \times r}$.   $\mathcal{T}_{RECAST}$  supports PEFT due to $a^r$'s small size even for large values of $K$.  Additionally, reducing  hyperparameter $r$ results in MC.  However, limiting $a^r$ to a vector severely restricts how much the basis matrix $B^{*r}_i$ can be adjusted for a task.  Thus, RECAST only reported improvements for very small sizes of $a^r$ (with less than 100 task-specific parameters in a ViT-S/16~\cite{tasnimRECAST2025}).

\method{} address this limitation, in part, by introducing a new hyperparameter $s$ to control the size of the mixing matrix, \ie, $A^{'rs}_i \in \mathbb{R}^{r \times s}$.   As the columns of $A$ is not set to a specific size, as in $\mathcal{T}$ in prior work, we must also adjust the dimensions of $B$ to ensure the correct number of weights are produced \ie, $B^{'r}_i \in \mathbb{R}^{u \times r}$, where $u=\frac{d_{\text{in}}* d_{\text{out}}}{s}$.   

A common way of adding additional expressivity is to add a non-linearity to a function (\eg, ReLU~\cite{pmlr-v15-glorot11a}). \citet{Plummer2020NeuralPA} explored adding non-nonlinearities to a small hypernetwork~\cite{DBLP:conf/iclr/HaDL17} that predicted the parameters $a^r_{i,j}$ for a transformation similar to \cref{eq:recast}, but found them ineffective as the final transformation is still linear. An alternative is to add a non-linearity to the predicted weight matrix $W_i$, \ie, $\mathcal{T} = \phi(W_i)$.  However, this acts as a hard constraint on the layer weights, which may hurt performance.  This is exemplified by the fact that most networks might apply weight decay to a layer as a regularizer instead of forcing the weights into a specific form. Thus, we add expressivity without overfitting by applying a constraint only to the mixing matrix, 
\ie, the \method{} transformation is:
\begin{equation}
 \mathcal{T}_{\method{}}(B^{'r}_i,A^{'rs}_i) =  B^{'r}_i \left(\sigma(A^{'rs}_i) \odot A^{'rs}_i\right)
\label{eq:\method{}}
\end{equation}
where $\sigma(\cdot)$ is a sigmoid and $\odot$ denotes element-wise multiplication.  Note that \cref{eq:\method{}} requires reshaping into the correct dimensions before it can be used, \ie, $W_i = \text{reshape}\left(\mathcal{T}_{\method{}}(B^{'r}_i,A^{'rs}_i), (d_{\text{out}}, d_{\text{in}})\right)$.  While the restrictions placed on $A^{'rs}_i$ has the same form as the SiLU activation function~\cite{ELFWING20183}, it acts more as a constraint on the weights rather than a nonlinearity as $A^{'rs}_i$ contains \emph{only} layer-specific tuneable parameters.  

Prior work has also unsuccessfully tried restrict the output of $A$ before, but these typically used strong constraints, \eg, forcing $a^r$ to sum to one~\cite{Plummer2020NeuralPA}.  We find even similar functions like ReLU resulted in a significant drop in performance as it would result in a large number of zero-value weights in the output. Additionally, the smoother function provided by adapting SiLU helps reduce overfitting without adding new hyperparameters to tune as would be required by other regularization techniques like weight decay.  The final result is a transformation that enables us to use a single hyperparameter $s$ to adapt the relative size of the frozen basis vectors $B^{'r}_i$ and the tuneable mixing matrix $A^{'rs}_i$ to the needs of a particular application rather than the hard-coded restrictive formulations of prior work.

\subsection{Retrofitting Pretrained Models with \method{}}
\label{sec:retrofitting}

In \cref{sec:reparameterization} we discussed how \method{} generates weights by combing shared basis vectors $B^{'r}_i$ and the tuneable mixing matrix $A^{'rs}_i$.  However, most off-the-shelf models do not provide these matrices.  Instead, we must retrofit a pretrained model to use our \method{} framework.  Following~\cite{tasnimRECAST2025,wang2025basis}, we initialize our weight matrices using a short preprocessing step.  Specifically, we begin by grouping together layers of the same module of consecutive layers that will share parameters (\eg, QKV attention or projection). Recall that in \cref{eq:\method{}} only $B^{'r}_i$ is shared across layers, whereas $A^{'rs}_i$ is learned per-layer. Then, we decompose the pretrained model by adopting Neural Mimicry~\cite{tasnimRECAST2025}, which learns to reconstruct the $N$ layer weights $W_{p_i}$ via:
\begin{equation}
\mathcal{L}_{mimicry} =  \sum^N_{i=1} \ell_{smL1}\left(\mathcal{T}_{\method{}}(B^{'r}_i,A^{'rs}_i) - W_{p_i}\right),
\label{eq:mimic}
\end{equation}
where $\ell_{smL1}$ refers to smooth-L1 loss~\cite{Girshick_2015_ICCV}.  

Note that \cref{eq:mimic} does not use any dataset samples as it simply retrofits a pretrained model into our framework.  Thus, this process has negligible computational requirements.  For example, in our experiments it completes in under a minute on a single GPU for ViTs~\cite{DBLP:journals/corr/abs-2010-11929} and $<30$ minutes for Llama models~\cite{li2023llama,dubey2024llama3}. We observe \cref{eq:mimic} alone is suitable when $s$ and $r$ are relatively large (\eg, resulting in similar parameters as the original model), but find learning $A^{'rs}_i$ and $B^{'r}_i$ under aggressive compression challenging.  

We make two changes to assist with the learning process in these settings where significant compression is desired.  First, we obtain a strong initialization for our target model $\mathcal{M}_{student}$ by using the top $r$ eigenvectors of a $M_{Teacher}$ model's $A^{'rs}_i$ and $B^{'r}_i$ matrices.  This $M_{Teacher}$ is simply a full parameter model created using \cref{eq:mimic} (\ie, it has the same parameter count as the pretrained model).  Second, we use $M_{Teacher}$ to guide the learning of $\mathcal{M}_{student}$ by using both KL divergence and MSE loss on the final predictions of $M_{Teacher}$ and $\mathcal{M}_{student}$ along with per-layer MSE feature matching.  During this compression stage we use just 2\% of ImageNet~\cite{5206848}, making it significantly more efficient than recent MC competitors who use the entire ImageNet dataset~\cite{Ahmed_2025_CVPR}.  See the supplementary for additional details.


\section{Experiments}
\label{sec: experiments}

\begin{table*}[t]
\centering
\footnotesize
\setlength{\tabcolsep}{3pt}
\caption{PEFT performance on VTAB-1K~\cite{zhai2019large} across 19 tasks grouped into Natural (7), Specialized (4), and Structured (8). \method{} achieves state-of-the-art overall accuracy while tuning 28\% fewer parameters than all baselines ($5\times10^{-3}\%$ vs.\ $7\times10^{-3}\%$ of the base model). It is particularly strong on Structured tasks, achieving the best result on the majority of benchmarks. Bold indicates best performance per task.}
\label{tab:vtab_results}
\begin{tabular}{l@{\hspace{3pt}}c@{\hspace{3pt}}c@{\hspace{3pt}}c@{\hspace{3pt}}c@{\hspace{3pt}}c@{\hspace{3pt}}c@{\hspace{3pt}}c@{\hspace{3pt}}|c|@{\hspace{3pt}}c@{\hspace{3pt}}c@{\hspace{3pt}}c@{\hspace{3pt}}c@{\hspace{3pt}}|c|@{\hspace{3pt}}c@{\hspace{3pt}}c@{\hspace{3pt}}c@{\hspace{3pt}}c@{\hspace{3pt}}c@{\hspace{3pt}}c@{\hspace{3pt}}c@{\hspace{3pt}}c@{\hspace{3pt}}|c|@{\hspace{3pt}}c}
\toprule
\multirow{2}{*}{\textbf{Method}}
    & \multicolumn{8}{c}{\cellcolor{blue!15}\textbf{Natural}}
    & \multicolumn{5}{c}{\cellcolor{green!15}\textbf{Specialized}}
    & \multicolumn{9}{c}{\cellcolor{orange!15}\textbf{Structured}}
    & \\
\cmidrule(lr){2-9} \cmidrule(lr){10-14} \cmidrule(lr){15-23}
& \cellcolor{blue!15}\rotatebox{90}{\textbf{CIFAR-100}}
& \cellcolor{blue!15}\rotatebox{90}{\textbf{Cltch101}}
& \cellcolor{blue!15}\rotatebox{90}{\textbf{DTD}}
& \cellcolor{blue!15}\rotatebox{90}{\textbf{Flowers102}}
& \cellcolor{blue!15}\rotatebox{90}{\textbf{Pets}}
& \cellcolor{blue!15}\rotatebox{90}{\textbf{SVHN}}
& \cellcolor{blue!15}\rotatebox{90}{\textbf{Sun397}}
& \cellcolor{blue!15}\textbf{Mean}
& \cellcolor{green!15}\rotatebox{90}{\textbf{Camelyon}}
& \cellcolor{green!15}\rotatebox{90}{\textbf{EuroSAT}}
& \cellcolor{green!15}\rotatebox{90}{\textbf{Resisc45}}
& \cellcolor{green!15}\rotatebox{90}{\textbf{Retinopathy}}
& \cellcolor{green!15}\textbf{Mean}
& \cellcolor{orange!15}\rotatebox{90}{\textbf{Clevr-Count}}
& \cellcolor{orange!15}\rotatebox{90}{\textbf{Clevr-Dist}}
& \cellcolor{orange!15}\rotatebox{90}{\textbf{DMLab}}
& \cellcolor{orange!15}\rotatebox{90}{\textbf{KITTI-Dist}}
& \cellcolor{orange!15}\rotatebox{90}{\textbf{dSpr-Loc}}
& \cellcolor{orange!15}\rotatebox{90}{\textbf{dSpr-Ori}}
& \cellcolor{orange!15}\rotatebox{90}{\textbf{sNORB-Azim}}
& \cellcolor{orange!15}\rotatebox{90}{\textbf{sNORB-Elev}}
& \cellcolor{orange!15}\textbf{Mean}
& \textbf{Overall} \\
\midrule
ViT-S/16~\cite{DBLP:journals/corr/abs-2010-11929}
    & \cellcolor{blue!15}42.1 & \cellcolor{blue!15}84.4 & \cellcolor{blue!15}62.6 & \cellcolor{blue!15}97.3 & \cellcolor{blue!15}87.2 & \cellcolor{blue!15}31.1 & \cellcolor{blue!15}43.8 & \cellcolor{blue!15}64.0
    & \cellcolor{green!15}74.8 & \cellcolor{green!15}87.8 & \cellcolor{green!15}75.2 & \cellcolor{green!15}70.8 & \cellcolor{green!15}77.1
    & \cellcolor{orange!15}34.9 & \cellcolor{orange!15}29.8 & \cellcolor{orange!15}34.8 & \cellcolor{orange!15}45.9 & \cellcolor{orange!15}14.5 & \cellcolor{orange!15}15.1 & \cellcolor{orange!15}11.1 & \cellcolor{orange!15}20.0 & \cellcolor{orange!15}25.7
    & 50.7 \\
\midrule
BoFT~\cite{liu2024boft}
    & \cellcolor{blue!15}34.4 & \cellcolor{blue!15}86.1 & \cellcolor{blue!15}60.5 & \cellcolor{blue!15}97.3 & \cellcolor{blue!15}82.2 & \cellcolor{blue!15}61.4 & \cellcolor{blue!15}35.8 & \cellcolor{blue!15}65.3
    & \cellcolor{green!15}73.5 & \cellcolor{green!15}89.6 & \cellcolor{green!15}73.3 & \cellcolor{green!15}68.4 & \cellcolor{green!15}76.1
    & \cellcolor{orange!15}31.2 & \cellcolor{orange!15}45.1 & \cellcolor{orange!15}36.0 & \cellcolor{orange!15}47.5 & \cellcolor{orange!15}22.6 & \cellcolor{orange!15}16.3 & \cellcolor{orange!15}12.6 & \cellcolor{orange!15}25.6 & \cellcolor{orange!15}29.6
    & 52.6 \\
DoRA~\cite{10.5555/3692070.3693369}
    & \cellcolor{blue!15}38.5 & \cellcolor{blue!15}86.9 & \cellcolor{blue!15}60.6 & \cellcolor{blue!15}97.2 & \cellcolor{blue!15}85.9 & \cellcolor{blue!15}52.7 & \cellcolor{blue!15}33.2 & \cellcolor{blue!15}65.0
    & \cellcolor{green!15}72.9 & \cellcolor{green!15}86.0 & \cellcolor{green!15}75.3 & \cellcolor{green!15}66.9 & \cellcolor{green!15}75.2
    & \cellcolor{orange!15}46.1 & \cellcolor{orange!15}45.7 & \cellcolor{orange!15}34.1 & \cellcolor{orange!15}41.6 & \cellcolor{orange!15}21.4 & \cellcolor{orange!15}21.9 & \cellcolor{orange!15}11.2 & \cellcolor{orange!15}25.1 & \cellcolor{orange!15}30.8
    & 52.8 \\
IA3~\cite{10.5555/3600270.3600412}
    & \cellcolor{blue!15}11.3 & \cellcolor{blue!15}24.5 & \cellcolor{blue!15}25.9 & \cellcolor{blue!15}41.1 & \cellcolor{blue!15}12.8 & \cellcolor{blue!15}34.7 & \cellcolor{blue!15}7.6 & \cellcolor{blue!15}22.5
    & \cellcolor{green!15}71.2 & \cellcolor{green!15}75.0 & \cellcolor{green!15}43.9 & \cellcolor{green!15}71.9 & \cellcolor{green!15}65.4
    & \cellcolor{orange!15}35.6 & \cellcolor{orange!15}50.9 & \cellcolor{orange!15}24.4 & \cellcolor{orange!15}39.5 & \cellcolor{orange!15}22.6 & \cellcolor{orange!15}10.4 & \cellcolor{orange!15}7.8 & \cellcolor{orange!15}18.2 & \cellcolor{orange!15}26.1
    & 33.1 \\
LoRA~\cite{hu2022lora}
    & \cellcolor{blue!15}52.8 & \cellcolor{blue!15}85.0 & \cellcolor{blue!15}64.3 & \cellcolor{blue!15}98.1 & \cellcolor{blue!15}88.4 & \cellcolor{blue!15}55.9 & \cellcolor{blue!15}45.4 & \cellcolor{blue!15}69.9
    & \cellcolor{green!15}75.6 & \cellcolor{green!15}90.4 & \cellcolor{green!15}79.0 & \cellcolor{green!15}69.3 & \cellcolor{green!15}78.5
    & \cellcolor{orange!15}45.0 & \cellcolor{orange!15}47.6 & \cellcolor{orange!15}39.5 & \cellcolor{orange!15}\textbf{52.7} & \cellcolor{orange!15}19.8 & \cellcolor{orange!15}23.1 & \cellcolor{orange!15}11.7 & \cellcolor{orange!15}18.2 & \cellcolor{orange!15}32.1
    & 55.8 \\
ReCAST~\cite{tasnimRECAST2025}
    & \cellcolor{blue!15}50.8 & \cellcolor{blue!15}85.1 & \cellcolor{blue!15}65.5 & \cellcolor{blue!15}98.6 & \cellcolor{blue!15}87.7 & \cellcolor{blue!15}45.1 & \cellcolor{blue!15}46.8 & \cellcolor{blue!15}68.5
    & \cellcolor{green!15}76.2 & \cellcolor{green!15}91.1 & \cellcolor{green!15}77.1 & \cellcolor{green!15}71.8 & \cellcolor{green!15}79.0
    & \cellcolor{orange!15}44.2 & \cellcolor{orange!15}42.1 & \cellcolor{orange!15}38.3 & \cellcolor{orange!15}51.3 & \cellcolor{orange!15}21.0 & \cellcolor{orange!15}24.3 & \cellcolor{orange!15}11.6 & \cellcolor{orange!15}24.0 & \cellcolor{orange!15}32.0
    & 55.3 \\
RoAD~\cite{liao2024in}
    & \cellcolor{blue!15}52.8 & \cellcolor{blue!15}85.8 & \cellcolor{blue!15}66.5 & \cellcolor{blue!15}\textbf{98.8} & \cellcolor{blue!15}88.9 & \cellcolor{blue!15}60.9 & \cellcolor{blue!15}46.8 & \cellcolor{blue!15}71.4
    & \cellcolor{green!15}\textbf{77.0} & \cellcolor{green!15}91.2 & \cellcolor{green!15}80.1 & \cellcolor{green!15}69.6 & \cellcolor{green!15}79.4
    & \cellcolor{orange!15}50.6 & \cellcolor{orange!15}49.3 & \cellcolor{orange!15}40.1 & \cellcolor{orange!15}51.2 & \cellcolor{orange!15}21.8 & \cellcolor{orange!15}\textbf{24.9} & \cellcolor{orange!15}11.7 & \cellcolor{orange!15}26.1 & \cellcolor{orange!15}34.4
    & 57.5 \\
RoSA~\cite{nikdan2024rosa}
    & \cellcolor{blue!15}42.4 & \cellcolor{blue!15}83.2 & \cellcolor{blue!15}61.8 & \cellcolor{blue!15}97.7 & \cellcolor{blue!15}87.2 & \cellcolor{blue!15}32.3 & \cellcolor{blue!15}43.6 & \cellcolor{blue!15}64.0
    & \cellcolor{green!15}76.2 & \cellcolor{green!15}88.2 & \cellcolor{green!15}72.6 & \cellcolor{green!15}69.8 & \cellcolor{green!15}76.7
    & \cellcolor{orange!15}35.4 & \cellcolor{orange!15}28.1 & \cellcolor{orange!15}34.5 & \cellcolor{orange!15}43.9 & \cellcolor{orange!15}14.3 & \cellcolor{orange!15}16.6 & \cellcolor{orange!15}11.1 & \cellcolor{orange!15}20.0 & \cellcolor{orange!15}25.4
    & 50.4 \\
SSF~\cite{Lian_2022_SSF}
    & \cellcolor{blue!15}\textbf{58.8} & \cellcolor{blue!15}\textbf{87.0} & \cellcolor{blue!15}\textbf{66.8} & \cellcolor{blue!15}\textbf{98.8} & \cellcolor{blue!15}\textbf{89.5} & \cellcolor{blue!15}\textbf{68.0} & \cellcolor{blue!15}47.0 & \cellcolor{blue!15}\textbf{73.7}
    & \cellcolor{green!15}76.0 & \cellcolor{green!15}91.4 & \cellcolor{green!15}\textbf{81.3} & \cellcolor{green!15}71.8 & \cellcolor{green!15}80.1
    & \cellcolor{orange!15}46.9 & \cellcolor{orange!15}42.5 & \cellcolor{orange!15}39.5 & \cellcolor{orange!15}50.9 & \cellcolor{orange!15}20.4 & \cellcolor{orange!15}24.4 & \cellcolor{orange!15}12.5 & \cellcolor{orange!15}24.9 & \cellcolor{orange!15}32.7
    & 57.8 \\
VBLoRA~\cite{li2024vblora}
    & \cellcolor{blue!15}50.9 & \cellcolor{blue!15}85.0 & \cellcolor{blue!15}64.6 & \cellcolor{blue!15}98.2 & \cellcolor{blue!15}87.9 & \cellcolor{blue!15}49.7 & \cellcolor{blue!15}44.9 & \cellcolor{blue!15}68.7
    & \cellcolor{green!15}74.2 & \cellcolor{green!15}90.2 & \cellcolor{green!15}77.1 & \cellcolor{green!15}69.4 & \cellcolor{green!15}77.7
    & \cellcolor{orange!15}46.5 & \cellcolor{orange!15}39.8 & \cellcolor{orange!15}37.2 & \cellcolor{orange!15}46.7 & \cellcolor{orange!15}19.2 & \cellcolor{orange!15}20.0 & \cellcolor{orange!15}11.8 & \cellcolor{orange!15}24.9 & \cellcolor{orange!15}30.7
    & 54.6 \\
SVFT~\cite{lingam2024svft}
    & \cellcolor{blue!15}50.5 & \cellcolor{blue!15}84.2 & \cellcolor{blue!15}64.4 & \cellcolor{blue!15}97.8 & \cellcolor{blue!15}88.5 & \cellcolor{blue!15}45.6 & \cellcolor{blue!15}39.8 & \cellcolor{blue!15}67.3
    & \cellcolor{green!15}73.4 & \cellcolor{green!15}90.1 & \cellcolor{green!15}75.6 & \cellcolor{green!15}68.5 & \cellcolor{green!15}76.9
    & \cellcolor{orange!15}41.2 & \cellcolor{orange!15}36.2 & \cellcolor{orange!15}37.8 & \cellcolor{orange!15}48.1 & \cellcolor{orange!15}19.4 & \cellcolor{orange!15}20.4 & \cellcolor{orange!15}11.2 & \cellcolor{orange!15}23.7 & \cellcolor{orange!15}29.7
    & 53.5 \\
\midrule
\rowcolor{yellow!30}
\method{} (ours)
    & \cellcolor{yellow!30}56.7 & \cellcolor{yellow!30}\textbf{87.0} & \cellcolor{yellow!30}66.5 & \cellcolor{yellow!30}\textbf{98.8} & \cellcolor{yellow!30}87.6 & \cellcolor{yellow!30}67.3 & \cellcolor{yellow!30}\textbf{47.7} & \cellcolor{yellow!30}73.0
    & \cellcolor{yellow!30}76.8 & \cellcolor{yellow!30}\textbf{91.6} & \cellcolor{yellow!30}79.5 & \cellcolor{yellow!30}\textbf{74.0} & \cellcolor{yellow!30}\textbf{80.4}
    & \cellcolor{yellow!30}\textbf{61.8} & \cellcolor{yellow!30}\textbf{51.6} & \cellcolor{yellow!30}\textbf{40.5} & \cellcolor{yellow!30}48.4 & \cellcolor{yellow!30}\textbf{23.6} & \cellcolor{yellow!30}24.6 & \cellcolor{yellow!30}\textbf{12.9} & \cellcolor{yellow!30}\textbf{27.8} & \cellcolor{yellow!30}\textbf{36.4}
    & \cellcolor{yellow!30}\textbf{59.2} \\
\bottomrule
\end{tabular}
\end{table*}

\noindent\textbf{Datasets.} We evaluate our approach on a diverse set of experimental settings to demonstrate its ability to generalize.  First, we use VTAB-1K~\cite{zhai2019large}, a benchmark of 19 diverse visual tasks spanning Natural (7 tasks),  Specialized (4 tasks including medical and satellite imagery), and Structured (8 tasks requiring geometric and relational reasoning). Its limited training budget (1K samples per task) and broad task diversity make it well-suited for assessing sample efficiency and cross-domain generalization.  Second, we use a set of six fine-grained benchmarks to measure the effect of larger training budgets: Flowers102~\cite{Nilsback08}, FGVC-Aircraft~\cite{maji13fine-grained},
MIT-Indoor67~\cite{5206537}, CIFAR-100/10~\cite{Krizhevsky09}, and CUB-200-2011~\cite{WahCUB_200_2011}.  Finally, as the first two sets of datasets focus on vision tasks (using ViTs~\cite{DBLP:journals/corr/abs-2010-11929}), we validate \method{}'s ability to generalize to LLMs with seven commonsense reasoning benchmarks: BoolQ~\cite{clark-etal-2019-boolq}, PIQA~\cite{bisk2020piqa}, SIQA~\cite{sap-etal-2019-social}, HellaSwag~\cite{zellers2019hellaswag}, WinoGrande~\cite{10.1145/3474381}, ARC-easy~\cite{clark2018thinksolvedquestionanswering}, ARC-challenge~\cite{clark2018thinksolvedquestionanswering}, and OBQA~\cite{mihaylov-etal-2018-suit}. We also report quantization results reported on the ImageNet-1K~\cite{5206848} dataset.
\smallskip

\noindent\textbf{Implementation Details.}
 All ViT~\cite{DBLP:journals/corr/abs-2010-11929} backbones are pretrained on ImageNet~\cite{5206848} and retrofitted using up to 1000 epochs ($\text{lr}=0.01$, Step Scheduler), followed by fine-tuning with 
AdamW~\cite{loshchilov2019adamw} for 100 epochs with early stopping. Following \cite{tasnimRECAST2025}, we apply layer-wise adapter parameter sharing across consecutive layer groups and structured binary masking for more extreme compression ratios.
For LLM experiments, we use models from the LLaMA family~\cite{dubey2024llama3}, adopting the experimental settings of~\cite{hu-etal-2023-llm} for CRISP coefficient fine-tuning. PEFT baseline results are taken directly from their respective papers, while compression baselines are reproduced using each method's official repository. All evaluations are performed using the \texttt{lm-evaluation-harness}~\cite{eval-harness} library under consistent hyperparameters across all methods. Additional details are provided in the supplementary.
\smallskip

\noindent\textbf{Baselines.}  We compare to methods from relevant prior work in both PEFT~\cite{10.5555/3600270.3600412,Lian_2022_SSF,hu2022lora,10.5555/3692070.3693369,lingam2024svft,li2024vblora,liao2024in,nikdan2024rosa,liu2024boft} and MC~\cite{wang2025basis,11072282,gonzalezmartinez2025balfbudgetedactivationawarelowrank,shen2025diversity,10.1007/978-3-031-73404-5_14,Ahmed_2025_CVPR}.  This includes methods like WeCoLoRA~\cite{grigore2024weight} whose adaptation based on LoRA makes it appear to support MC and PEFT at first glace.  We also compare to the closely related RECAST~\cite{tasnimRECAST2025} approach that was introduced as a PEFT method, but can naively support MC.  Finally, we provide a simple baseline that uses a SVD decomposition to support MC by retaining only the vectors $U$ and $V$ related to the top-k eigenvalues $\Sigma$, and then performs PEFT by updating $\Sigma$ while freezing $U$ and $V$.

\subsection{Isolated PEFT or MC Results}

\textbf{PEFT Only.}
\cref{tab:vtab_results} reports that \method{} boosts PEFT performance on VTAB-1K~\cite{zhai2019large} by 1.5\% over the state-of-the-art while using 28\% fewer trainable parameters. The largest gains appear on Structured tasks, where \method{} leads by 2\% on average and achieves best accuracy on most individual benchmarks.  In contrast methods like LoRA~\cite{hu2022lora} and RECAST~\cite{tasnimRECAST2025} plateau or slightly degrade here, suggesting that their linear updates struggle to capture the more complex task-specific structure these benchmarks demand. On Natural and Specialized tasks, \method{} remains competitive with SSF~\cite{Lian_2022_SSF} and RoAD~\cite{liao2024in} despite the smaller budget, indicating broad generalization across task types. 

\begin{figure}[t]
    \centering
    \includegraphics[width=\linewidth]{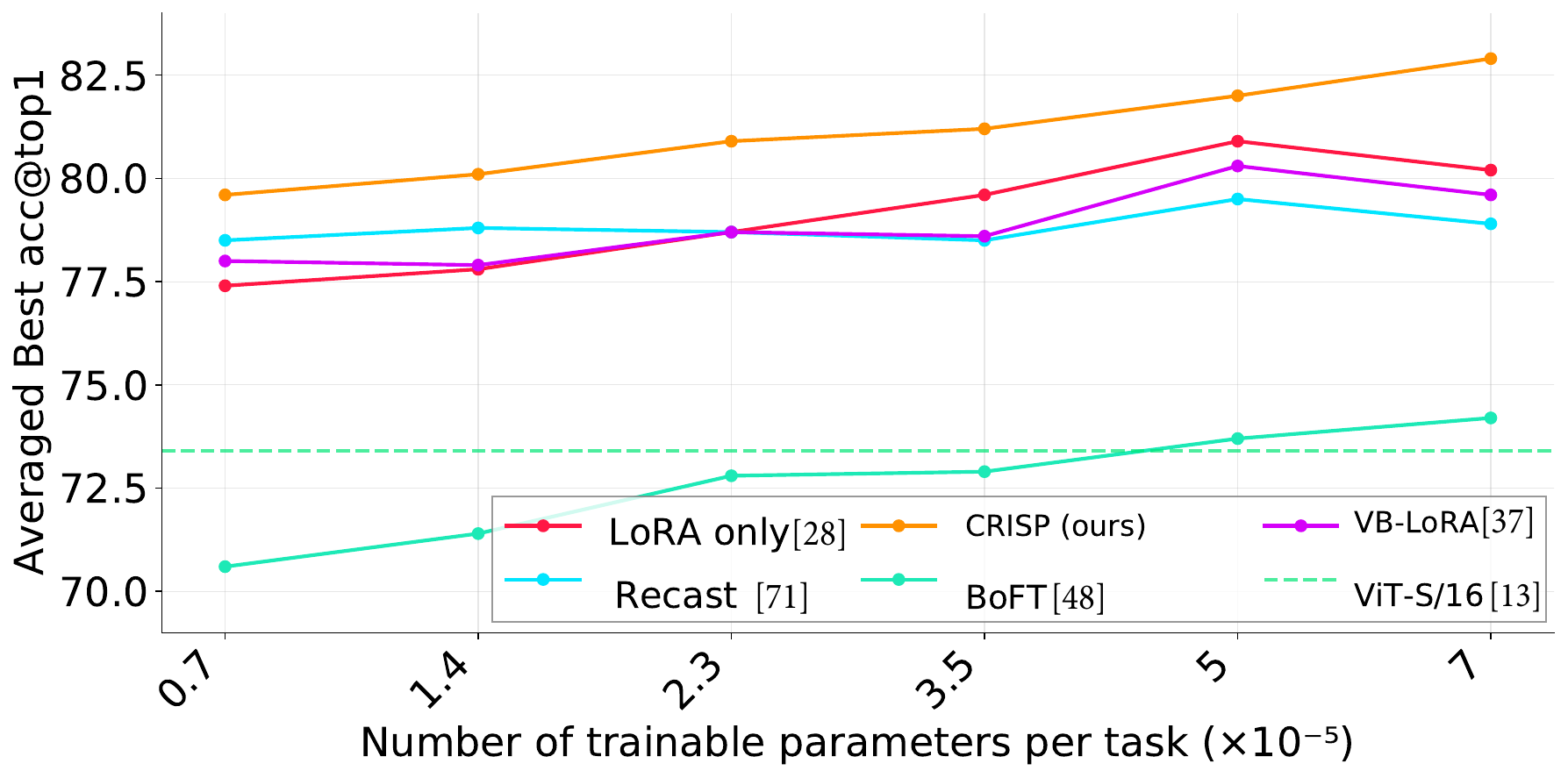}
    \vspace{-4mm}
    \caption{PEFT performance using a ViT-S/16 across a range of trainable parameter budgets averaged over three datasets: FGVC-Aircraft~\cite{maji13fine-grained}, CIFAR-100~\cite{Krizhevsky09} and CUB-200-2011~\cite{WahCUB_200_2011}. \method{} consistently outperforms prior work in all settings.}
    \label{fig:param_scaling}
\end{figure}

\cref{fig:param_scaling} shows \method{} outperforms prior work across a range of parameter budgets. Notably, closely related method RECAST's performance plateau's, whereas \method{} continues to improve as the budget increases. \method{} also matches LoRA's peak accuracy with little over half the parameters. These results highlight the benefits from our improved weight generation approach defined in \cref{eq:\method{}}.
\smallskip

\begin{table*}[t]
\centering
\caption{ViT-B/16~\cite{DBLP:journals/corr/abs-2010-11929} compression at 50\% parameter reduction across six fine-grained benchmarks.
\textbf{Upper:} post-compression accuracy with classifier-only adaptation; \method{} leads all
baselines using only 2\% of ImageNet-1K~\cite{5206848} for distillation. \textbf{Lower:} compressed backbones
used as initialization for PEFT; \method{} achieves state-of-the-art, outperforming the best
pruning+PEFT combination by about 1 point and RECAST with coefficient tuning by 5 points,
suggesting that compression quality directly bounds downstream task adaptability.}

\label{tab:taskwise}
\begin{tabular}{lcccccccc}
\toprule
\textbf{Compression} & \textbf{PEFT} & \textbf{Flowers} & \textbf{Aircraft} & \textbf{Scene} & \textbf{CFR100} & \textbf{CFR10} & \textbf{Birds} & \textbf{Avg} \\
\midrule
\rowcolor{gray!20}ViT-B/16~\cite{DBLP:journals/corr/abs-2010-11929}
    & -- & 96.7 & 70.9 & 84.5 & 76.3 & 97.0 & 84.6 & 85.0 \\
\midrule
\multicolumn{9}{l}{\cellcolor{red!30}\textbf{Compressed Performance}} \\
\rowcolor{red!15} SVD
    & -- & 81.8 & 43.1 & 65.6  & 55.4 & 83.0 & 39.3 & 61.3 \\
\rowcolor{red!15} Isomorph. Prune.~\cite{10.1007/978-3-031-73404-5_14}
    & -- & 94.7 & 61.5 & 77.6  & 76.3 & \textbf{93.2} & 70.0 & 78.8 \\
\rowcolor{red!15} RDHP~\cite{11072282}
    & -- & 95.3 & 63.8 & 74.7  & \textbf{78.1} & 89.2 & 61.7 & 77.1 \\
\rowcolor{red!15} DGMR~\cite{shen2025diversity}
    & -- & 96.4 & 71.6 & 79.8  & 76.9 & 92.1 & 74.3 & 81.9 \\
\rowcolor{red!15} BALF~\cite{gonzalezmartinez2025balfbudgetedactivationawarelowrank}
    & -- & 94.3 & 68.3 & 76.3  & 69.7 & 89.8 & 70.3 & 78.1 \\
\rowcolor{red!15} WeCoLoRA~\cite{grigore2024weight}
    & -- & 90.8 & 50.7 & 63.9  & 65.5 & 81.1 & 45.6 & 66.2 \\
\rowcolor{red!15} RECAST~\cite{tasnimRECAST2025}
    & -- & 92.1 & 71.8 & 77.4  & 71.5 & 91.8 & 71.0 & 79.3 \\
\rowcolor{red!40} \method{} (ours)
    & -- & \textbf{96.8} & \textbf{76.3} & \textbf{81.3}  & 76.2 & 92.1 & \textbf{77.5} & \textbf{83.3} \\
\midrule
\multicolumn{9}{l}{\cellcolor{blue!30}\textbf{Compressed + PEFT Performance}} \\
\rowcolor{blue!15} SVD
    & Eigenvalues & 85.6 & 56.9 & 68.1 & 71.5 & 92.9 & 58.5 & 72.3 \\
\rowcolor{blue!15} Isomorph. Prune.~\cite{10.1007/978-3-031-73404-5_14}
    & LoRA~\cite{hu2022lora} & 98.4 & 90.8 & 78.0 & 84.7 & 96.7 & 73.4 & 87.0 \\
\rowcolor{blue!15} Isomorph. Prune.~\cite{10.1007/978-3-031-73404-5_14}
    & SSF~\cite{Lian_2022_SSF}  & 98.2 & 86.2 & 81.7 & 85.6 & \textbf{97.8} & 77.1 & 87.7 \\
\rowcolor{blue!15} Isomorph. Prune.~\cite{10.1007/978-3-031-73404-5_14}
    & SVFT~\cite{lingam2024svft} & 98.2 & 89.5 & 80.6 & 84.4 & 97.0 & 74.2 & 87.3 \\

\rowcolor{blue!15} RDHP~\cite{11072282}
    & LoRA~\cite{hu2022lora} & 97.8 & \textbf{91.5} & 79.8 & 84.8 & 97.0 & 72.6 & 87.1 \\
\rowcolor{blue!15} RDHP~\cite{11072282}
    & SSF~\cite{Lian_2022_SSF}  & 98.0 & 85.0 & 77.6 & 83.9 & 96.5 & 71.0 & 85.3 \\
\rowcolor{blue!15} RDHP~\cite{11072282}
    & SVFT~\cite{lingam2024svft} & 98.2 & 89.9 & 77.6 & 83.1 & 97.1 & 72.6 & 86.4 \\
\rowcolor{blue!15} DGMR~\cite{shen2025diversity}
    & LoRA~\cite{hu2022lora} & 99.0 & 91.1 & 77.2 & 86.1 & 93.7 & 75.5 & 87.1 \\
\rowcolor{blue!15} DGMR~\cite{shen2025diversity}
    & SSF~\cite{Lian_2022_SSF}  & 98.8 & 87.2 & \textbf{82.4} & 84.8 & 97.2 & 76.8 & 87.9 \\
\rowcolor{blue!15} DGMR~\cite{shen2025diversity}
    & SVFT~\cite{lingam2024svft} & 98.8 & 91.2 & 77.4 & 82.8 & 96.5 & 74.2 & 86.8 \\
\rowcolor{blue!15} WeCoLoRA~\cite{grigore2024weight}
    & WeCoLoRA~\cite{grigore2024weight} &  3.3 & 20.9 &  4.0 & 19.3 & 10.1 &  6.8 & 10.7 \\
\rowcolor{blue!15} WeCoLoRA~\cite{grigore2024weight}
    & SSF~\cite{Lian_2022_SSF}  & 98.2  & 80.2 & 73.3  & 77.6 & 94.3 & 69.9 & 82.2 \\
\rowcolor{blue!15} WeCoLoRA~\cite{grigore2024weight}
    & SVFT~\cite{lingam2024svft} & 97.4  & 73.7 & 72.4  & 75.1 & 92.6 & 65.0 & 79.3 \\
\rowcolor{blue!15} RECAST~\cite{tasnimRECAST2025}
    & RECAST~\cite{tasnimRECAST2025}        & 96.8 & 77.3 & 79.3 & 78.5 & 95.3 & 74.9 & 83.7 \\
\rowcolor{blue!40} \method{} (ours)
    & \method{} (ours) & \textbf{99.0} & {89.5} & {81.8} & \textbf{86.2} & 97.4 & \textbf{79.1} & \textbf{88.8} \\
\bottomrule
\end{tabular}
\end{table*}

\begin{table*}[t]
\centering
\caption{LLaMA3.2-1B~\cite{dubey2024llama3} compression at 30\% parameter reduction across seven commonsense reasoning benchmarks. \method{}'s 3\% average gain shows it can perform well on LLMs, demonstrating an ability generalize across architectures we found many prior works lack.}
\label{tab:compression_llama}
\begin{tabular}{l@{\hspace{6pt}}c@{\hspace{6pt}}c@{\hspace{6pt}}c@{\hspace{6pt}}c@{\hspace{6pt}}c@{\hspace{6pt}}c@{\hspace{6pt}}c@{\hspace{6pt}}c}
\toprule
& {BQ} & {PIQ} & {Hell.} & {Wino} & {ARC-e} & {ARC-c} & {OBQ} & {Avg.} \\
\midrule
\rowcolor{gray!20} LLaMA3.2-1B~\cite{dubey2024llama3}  & 63.9 & 74.4 & 47.7 & 60.0 & 65.4 & 31.3 & 26.4 & 52.7 \\
\midrule
\rowcolor{red!15} SVD & 38.1 & 53.2 & 26.0 & 47.9 & 26.1 & \textbf{21.5} & 15.6 & 32.6 \\
\rowcolor{red!15} Basis-Sharing~\cite{wang2025basis} & 38.2 & 54.9 & 26.5 & 50.5 & 26.9 & 20.2 & 14.6 & 33.1 \\
\rowcolor{red!15} PruneNet~\cite{sengupta2025you} & 51.6 & 54.7 & 26.3 & 48.6 & 28.6 & 17.8 & \textbf{16.8} & 34.9 \\
\rowcolor{red!15} DFJR~\cite{DBLP:journals/corr/abs-2402-16319}  & 37.8 & 53.7 & 26.1 & 49.6 & 26.1 & 19.3 & 14.2 & 32.4 \\
\rowcolor{red!40} \method{} (ours)  & \textbf{60.2} & \textbf{57.5} & \textbf{27.7} & \textbf{50.6} & \textbf{34.6} & 20.1 & 15.2 & \textbf{38.0} \\
\bottomrule
\end{tabular}
\end{table*}

\noindent\textbf{MC Only.} \cref{tab:taskwise}(Upper) reports performance at 50\% parameter reduction (86M $\to$ 44M) using a ViT-B/16, where \method{} outperforms the state-of-the-art by 1.5\%.  Notably, prior compression methods such as DGMR~\cite{shen2025diversity}, RDHP~\cite{11072282}, and Isomorphic Pruning~\cite{10.1007/978-3-031-73404-5_14} use the full ImageNet-1K~\cite{5206848} dataset for distillation and recovery, whereas \method{} achieves superior performance using only 2\% of the same dataset, underscoring meaningful gains in data efficiency over prior work. \cref{tab:compression_llama} shows our approach generalizes to LLMs, where we outperform prior work by 3\% (additional experiments such as LLMs with PEFT are in the supplementary due to space constraints). These results demonstrate an ability to generalize across architectures we found many other methods lack.

\begin{figure}[t]
    \centering
    \includegraphics[width=.99\linewidth]{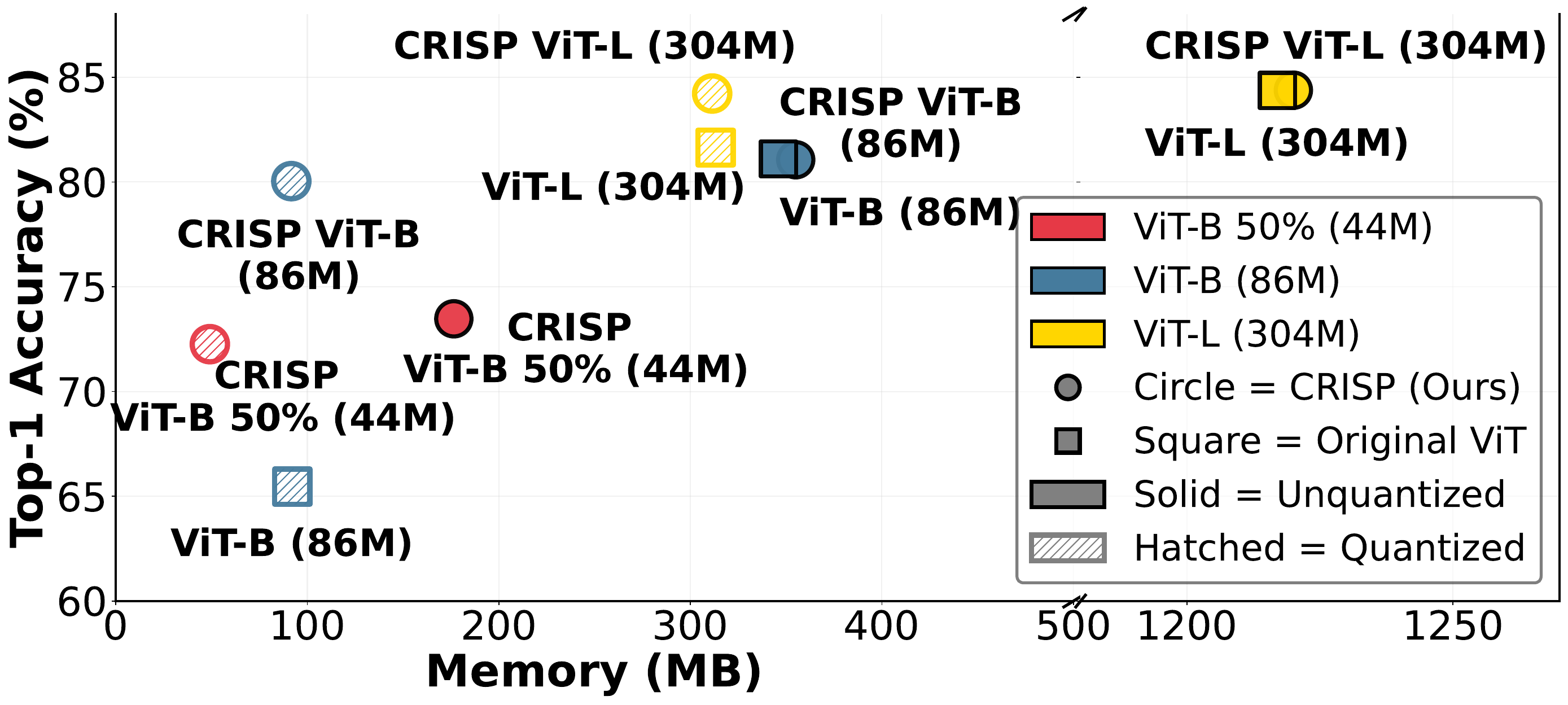}
    \caption{Comparing ImageNet~\cite{5206848} performance with and without 8-bit PTQ~\cite{wu2020integerquantizationdeeplearning} compression.  We find \method{} accurately reproduces the original model's performance while also demonstrating effective compositionality with other compression techniques.}
    \label{fig:quant_results}
\end{figure}

\cref{fig:quant_results} compares performance and memory utilization on ImageNet (the dataset our ViTs were pretrained on) of \method{} alone and in combination with orthogonal compression techniques like post-training quantization (PTQ)~\cite{wu2020integerquantizationdeeplearning}.  We make two primary observations on these results.  First, retrofitting these ViTs into our \method{} framework makes negligible difference in their pretraining dataset's performance, demonstrating that we accurately replicated the original model's performance.  Second, \method{} can be effectively combined with methods like PTQ for additional memory savings while outperforming either method alone.

\subsection{Combination PEFT and MC Results}
\label{sec:compression_results}
\cref{tab:taskwise}(Lower) reports performance of combinations of PEFT and MC that result in a total of $\sim$45M parameters (considering contributions of both PEFT and MC) for a model that originally contained 86M parameters. \method{} outperforms combinations of prior work by 1\% while also being simpler to implement due to our single unified framework. This gain increases to 5\% when restricting comparison to methods like RECAST~\cite{tasnimRECAST2025} that perform both PEFT and MC within the same framework.  Although MC method WeCoLoRA~\cite{grigore2024weight} is based on LoRA, we find using these matrices for PEFT collapses performance.  This suggests that the decomposition learned by WeCoLoRA is very sensitive to use directly.  \cref{tab:taskwise}(Lower) shows combining WeCoLoRA with feature-scaling adapters like SSF~\cite{Lian_2022_SSF} and SVFT~\cite{lingam2024svft} partially sidesteps this issue by operating on activations rather than weights.  However, \method{}'s unified framework where compression and adaptation are explicitly designed to coexist still outperforms it by 6.5\%.

\subsection{Model Analysis}
\label{sec:ablation}
\noindent\textbf{Regularization Methods.}
\cref{tab:reg_avg} compares regularization strategies applied to the mixer matrix $A^{'rs}_i$. Explicit techniques such as L2, orthogonal, and spectral normalization all provide modest gains over no regularization, but require an additional hyperparameter to tune. Activation-based constraints offer a parameter-free alternative: GELU and SiLU match or exceed explicit regularization, while ReLU collapses performance by zeroing out negative weights and producing overly sparse mixer matrices. In contrast, 
SiLU's smooth gating as defined in \cref{eq:\method{}} provides built-in regularization without introducing additional hyperparameters.
\smallskip

\begin{table}[t]
\centering
\caption{Ablation on the regularization strategy for the mixer matrix $A^{rs}_i$ using ViT-S/16~\cite{DBLP:journals/corr/abs-2010-11929} averaged over CUB-Birds, CIFAR-100, and FGVC-Aircraft. 
\textbf{(a) Explicit regularization} provides modest 
gains but requires additional hyperparameters. 
\textbf{(b) Activation-based constraints} offer a parameter-free alternative, where \method{}'s use of SiLU's smooth gating matches explicit regularization without introducing new hyperparameters. }
\label{tab:reg_avg}
\begin{tabular}{lcccc}
\hline
\textbf{Type} & \textbf{Birds} & \textbf{CIF100} & \textbf{ACraft} & \textbf{Av.} \\
\hline
No regularization & 81.5 & 86.0 & 77.9 & {81.8} \\
L2-Norm 
& 82.6 & 86.5 & 77.0 & {82.0} \\
Orthogonal 
& 84.2 & 86.9 & 76.1 & {82.3} \\
Spectral Norm & 83.5 & 85.7 & 77.4 & 82.1 \\
\midrule
ReLU  & 58.7 & 64.0 & 61.9 & {61.5} \\
GELU  & 80.1 & 85.9 & 79.0 & {81.6} \\
SiLU  & 81.5 & 86.2 & 79.1 & {82.2} \\
\hline
\end{tabular}
\end{table}

\begin{figure}[t]
    \centering
    \includegraphics[width=\linewidth]{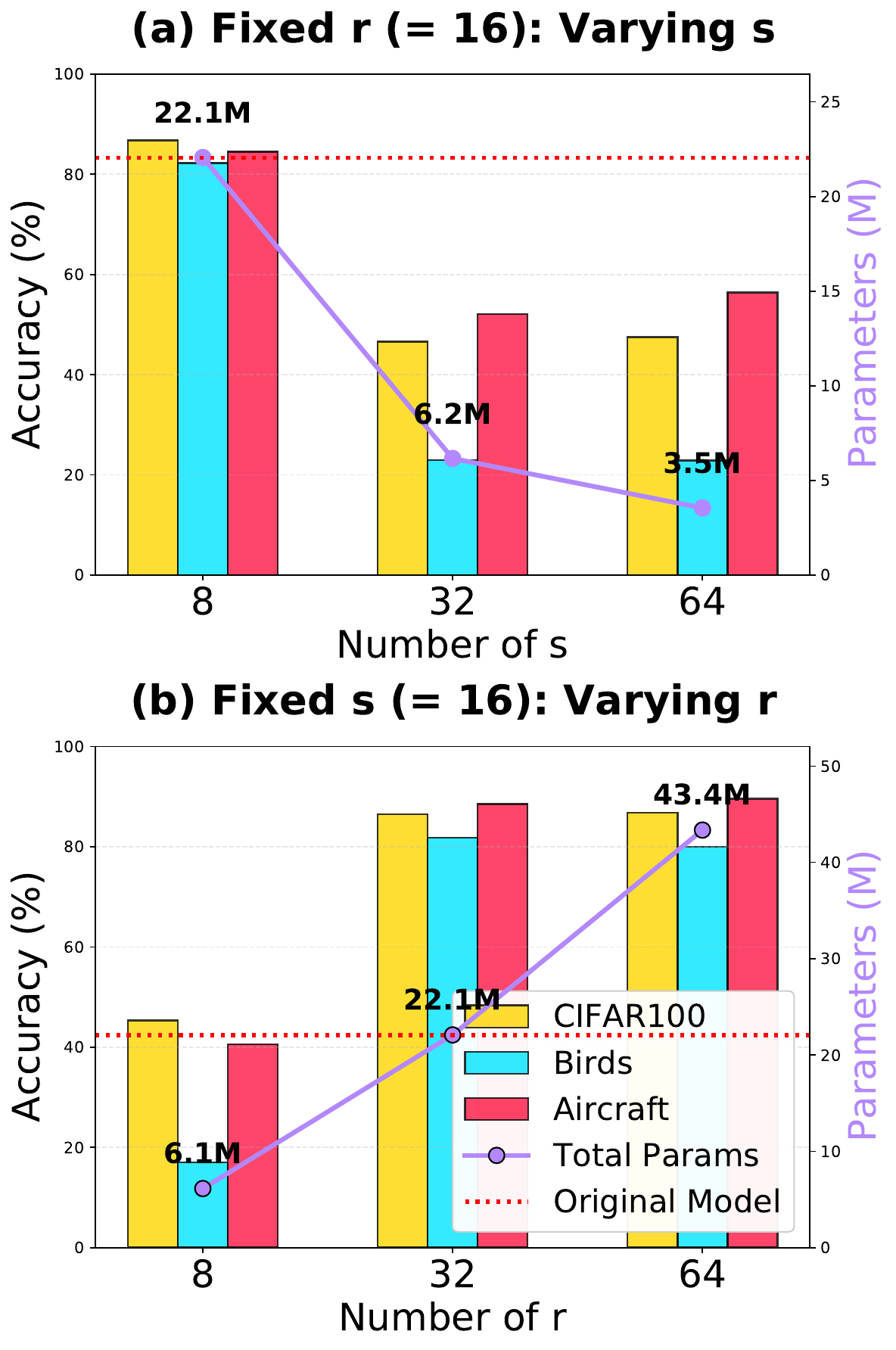}
    \vspace{-6mm}
    \caption{Impact of mixer matrix dimensions on model capacity and performance. \textbf{(a)} Fixed columns ($s=16$): Increasing rows reduces parameters but collapses accuracy. \textbf{(b)} Fixed rows ($r=16$): Increasing columns scales capacity and recovers performance. Results across CIFAR-100, CUB-Birds, and FGVC-Aircraft demonstrate that basis capacity (columns) is the dominant factor for maintaining model quality, while coefficient expressivity (rows) plays a secondary role. Red dotted line: original model.}

    \label{fig:mixer_ablation}
\end{figure}

\paragraph{Mixing Matrix Dimensions.}
\cref{fig:mixer_ablation} plots accuracy and total parameter count as $r$ and $s$ are varied independently, revealing a clear asymmetry between the two dimensions of $A^{'rs}_i$.
In \cref{fig:mixer_ablation}(b), \textit{fixing $s{=}16$ (columns) and increasing $r$ (rows)} from 8 to 64 shrinks the model from 22.1M to 3.5M parameters and causes accuracy to collapse across all three benchmarks, despite more rows providing greater per-layer expressivity.
The inverse experiment in \cref{fig:mixer_ablation}(a), \textit{fixing $r{=}16$ (rows) and increasing $s$ (columns)} from 8 to 32, restores both parameter count (6.1M to 22.1M) and strong accuracy, while further doubling to $s{=}64$ (43.4M) yields only marginal change. At matched budgets of $\sim$6M parameters, the column-starved configuration $(r{=}16, s{=}8)$ substantially underperforms the row-starved one $(r{=}32, s{=}16)$, confirming that
\textit{$s$, not $r$, is the primary driver of performance}.
This is consistent with \cref{eq:\method{}}: $s$ directly controls the dimensionality of the shared basis $B^{'r}_i$, and no amount of per-layer mixing via $A^{'rs}_i$ can compensate for an undercapacitated basis bank.

\subsection{Computational Efficiency Analysis}
\cref{tab:efficiency_comparison} compares training throughput, inference throughput, latency,
and peak memory under matched parameter budgets ($\sim$150K--240K) on the same ViT-B/16 backbone.
Adapters (and coefficients) are not merged into the backbone for any method; merging would improve inference uniformly across the board. 
\smallskip

\noindent\textbf{Training.}
\method{} achieves 163 samples/s training throughput, on par with LoRA (165), DoRA (178), and SSF (167), and substantially ahead of methods with more expensive forward passes such as BoFT (74).
Despite both \method{} and RECAST materializing weights from shared bases at each step, their training throughput differs by nearly 50$\times$ (163 vs.\ 3.3 samples/s). The key distinction lies in how weights are generated: RECAST averages over $K$ coefficient vectors per layer, requiring $K$ separate matrix multiplications in every forward and backward pass; \method{} performs a single matrix multiply with element-wise SiLU gating, keeping its cost constant regardless of basis size. This makes RECAST disproportionately expensive at training time
despite its small per-call generation overhead.
\smallskip

\noindent\textbf{Inference.}
\method{} achieves 657 samples/s inference throughput (1.52 ms latency), second only to DoRA (726 samples/s) and ahead of all other methods including IA3, SSF, and LoRA. Notably, \method{} achieves this despite materializing weights from shared bases at each step a cost that RECAST also incurs but handles far less efficiently: RECAST's $K$-fold template
averaging reduces its inference throughput to just 7 samples/s, nearly 100$\times$ slower, without caching. In terms of memory, \method{} and RECAST occupy a higher tier ($\sim$14 GB) than purely additive methods like LoRA and DoRA (4--5 GB) due to basis storage, yet \method{} delivers
competitive throughput that additive methods at this memory cost cannot match.
Overall, \method{} is the only weight-materializing method that matches top-tier PEFT methods in inference speed, while delivering superior compression quality and task adaptability.

\begin{table}[t]
\centering
\caption{Efficiency using a NVIDIA L40S for task adaptation on ViT-B/16~\cite{DBLP:journals/corr/abs-2010-11929} backbone under matched finetuning parameter budgets ($\sim$150K--240K trainable parameters, excluding head).}
\label{tab:efficiency_comparison}
\setlength{\tabcolsep}{4pt}
\begin{tabular}{lrrrr}
\toprule
\textbf{Method} & \textbf{Tr.Thr.} & \textbf{In.Thr.} & \textbf{Lat.} & \textbf{Mem.} \\
 & \scriptsize{samp/s} & \scriptsize{samp/s} & \scriptsize{ms} & \scriptsize{GB} \\
\midrule
LoRA~\cite{hu2022lora}      & 165.2 &  575 &  1.74 &  4.58 \\
DoRA~\cite{10.5555/3692070.3693369}      & 178.3 &  726 &  1.38 &  3.98 \\
BoFT~\cite{liu2024boft}      &  74.7 &  155 &  6.43 &  5.24 \\
IA3~\cite{10.5555/3600270.3600412}       & 156.8 &  629 &  1.59 &  6.87 \\
RoAD~\cite{liao2024in}      & 131.0 &  479 &  2.09 &  8.40 \\
SVFT~\cite{lingam2024svft}     & 134.2 &  418 &  2.39 &  5.15 \\
VBLoRA~\cite{li2024vblora}   & 161.1 &  513 &  1.95 &  7.72 \\
SSF~\cite{Lian_2022_SSF}     & 167.0 &  580 &  1.72 & 19.66 \\
RECAST~\cite{tasnimRECAST2025}   &   3.3 &    7 & 148.27 & 14.67 \\
\midrule
\rowcolor{blue!40}\method{} (ours) & 163.0 &  657 &  1.52 & 13.88 \\
\bottomrule
\end{tabular}
\end{table}

\section{Conclusion and Future Work}
\label{sec: discussion}
We introduced CRISP, a unified parameter recombination framework that seamlessly integrates model compression and parameter-efficient fine-tuning through a single architectural paradigm. Our coefficient-gated transformation decomposes pretrained weights into shared basis matrices and compact mixing coefficients, enabling flexible control over the expressivity-efficiency trade-off via hyperparameter $s$ while providing built-in regularization through SiLU-inspired gating. Our results demonstrate that unification need not compromise specialization: CRISP achieves a 1.5\% gain on PEFT while simultaneously outperforming similar PR methods like RECAST by 5\%. The broader implications extend beyond performance metrics, CRISP challenges the prevailing paradigm of developing isolated solutions for PR tasks and suggests that principled unification offers a more sustainable path forward. 

Future work might explore several directions: fully extending CRISP to language models where compression and adaptation demands are even more acute; investigating learned hyperparameter selection to automate capacity-efficiency trade-off; and combining CRISP with orthogonal techniques like quantization and pruning to compound gains. As foundation models continue scaling,  unified frameworks that elegantly balance multiple deployment constraints will become essential infrastructure. 
\clearpage
\setcounter{page}{1}
\maketitlesupplementary
\section{\method{} Algorithms}

\method{} consists of three core stages: weight reparameterization via the forward pass (Alg.~\ref{alg:crisp_forward}), neural mimicry retrofitting (Alg.~\ref{alg:neural_mimicry}), and task adaptation via PEFT (Alg.~\ref{alg:peft}). The compression stage differs between ViT and LLaMA architectures and is described separately below. The complete joint MC+PEFT pipeline is presented in Alg.~\ref{alg:joint}.

\cref{alg:crisp_forward} describes the forward pass using the 
coefficient-gated transformation $\mathcal{T}_{\text{CRISP}}(B_i', 
A_i'^{rs}) = B_i'(\sigma(A_i'^{rs}) \odot A_i'^{rs})$ introduced in 
Eq.~4 of the main paper. 
The retrofitting process in \cref{alg:neural_mimicry} implements the initial neural mimicry stage (Sec.\ 3.2), which decomposes pretrained weights into basis-mixer pairs through smooth-L1 reconstruction loss without requiring any dataset samples. 
\cref{alg:peft} demonstrates CRISP's PEFT capability, where only the 
lightweight mixer matrices $\{A_i'^{rs}\}$ are updated while basis 
matrices $\{B_i'\}$ remain frozen, enabling task adaptation with fewer than 200 trainable parameters per layer in some experiments 
(Tab.~1, main paper). 

\begin{algorithm}
\caption{CRISP Forward Pass}
\label{alg:crisp_forward}
\begin{algorithmic}[1]
\Require Input $\mathbf{x} \in \mathbb{R}^{n \times d_{\text{in}}}$, Basis $B_i' \in \mathbb{R}^{u \times r}$, Mixer $A_i'^{rs} \in \mathbb{R}^{r \times s}$, bias $\mathbf{b} \in \mathbb{R}^{d_{\text{out}}}$
\Ensure Output $\mathbf{y} \in \mathbb{R}^{n \times d_{\text{out}}}$
\State $\tilde{A}_i'^{rs} \gets \sigma(A_i'^{rs}) \odot A_i'^{rs}$ where $\sigma(\cdot)$ is sigmoid
\State $W_i \gets \text{reshape}(B_i' \tilde{A}_i'^{rs}, (d_{\text{out}}, d_{\text{in}}))$ where $u = \frac{d_{\text{in}} \cdot d_{\text{out}}}{s}$
\State $\mathbf{y} \gets \mathbf{x} W_i^T + \mathbf{b}$
\State \Return $\mathbf{y}$
\end{algorithmic}
\end{algorithm}

\begin{algorithm}
\caption{Neural Mimicry Initialization}
\label{alg:neural_mimicry}
\begin{algorithmic}[1]
\Require Pretrained weights $\{W_p^i\}_{i=1}^N$, layer groups $\mathcal{G}$, hyperparameters $r, s$
\Ensure Shared bases $\{B_i'\}$, per-layer mixers $\{A_i'^{rs}\}$
\For{each group $g \in \mathcal{G}$}
    \State Initialize shared basis $B_g' \sim \mathcal{N}(0, 0.01)$ of size $u \times r$
    \For{each layer $i \in g$}
        \State Initialize mixer $A_i'^{rs} \sim \mathcal{N}(0, 0.01)$ of size $r \times s$
    \EndFor
\EndFor
\While{not converged}
    \State $\mathcal{L}_{\text{mimicry}} \gets \sum_{i=1}^N \ell_{\text{smL1}}(\mathcal{T}_{\text{CRISP}}(B_i', A_i'^{rs}) - W_p^i)$
    \State Update $\{B_i'\}, \{A_i'^{rs}\}$ via gradient descent on $\mathcal{L}_{\text{mimicry}}$
\EndWhile
\State \Return $\{B_i'\}, \{A_i'^{rs}\}$
\end{algorithmic}
\end{algorithm}
\begin{algorithm}
\caption{CRISP PEFT Adaptation}
\label{alg:peft}
\begin{algorithmic}[1]
\Require Downstream dataset $\mathcal{D}$, compressed model with frozen bases $\{B_i'\}$, trainable mixers $\{A_i'^{rs}\}$, learning rate $\eta$
\Ensure Task-adapted model
\For{$(\mathbf{x}, y) \in \mathcal{D}$}
    \State $\hat{y} \gets \text{Forward}(\mathbf{x}; \{B_i'\}, \{A_i'^{rs}\})$ \Comment{Alg.~\ref{alg:crisp_forward}}
    \State $\mathcal{L}_{\text{task}} \gets \text{CrossEntropy}(\hat{y}, y)$
    \State $\{A_i'^{rs}\} \gets \{A_i'^{rs}\} - \eta \nabla_{\{A_i'^{rs}\}} \mathcal{L}_{\text{task}}$ \Comment{Freeze $\{B_i'\}$}
\EndFor
\State \Return adapted $\{A_i'^{rs}\}$
\end{algorithmic}
\end{algorithm}

\subsection{ViT Compression}
\label{sec:vit_compression_alg}
For ViT models, compression is performed via distillation 
using a full-parameter CRISP teacher model trained on only 2\% of 
ImageNet-1K~\cite{5206848}. \cref{alg:compression} initializes the 
student model using the top-$r$ eigenvectors of the teacher's basis 
matrices and optimizes a weighted combination of KL divergence on output logits and per-layer MSE feature matching 
(Tab.~2, main paper; Supp.~\cref{tab: taskwise,tab: compression_ablate}).
\begin{algorithm}
\caption{CRISP-ViT Compression via Distillation}
\label{alg:compression}
\begin{algorithmic}[1]
\Require Teacher model $\mathcal{M}_{\text{teacher}}$ (full CRISP from 
    Alg.~\ref{alg:neural_mimicry}), target compression $r_{\text{target}}, 
    s_{\text{target}}$, distillation dataset $\mathcal{D}_{\text{dist}}$ 
    (2\% ImageNet), loss weights $\lambda_{\text{KL}}, \lambda_{\text{feat}}$
\Ensure Compressed student model $\mathcal{M}_{\text{student}}$
\State Initialize $\{B_{\text{student}}'\}, \{A_{\text{student}}'^{rs}\}$ 
    with top $r_{\text{target}}$ eigenvectors from $\mathcal{M}_{\text{teacher}}$
\For{$(\mathbf{x}, y) \in \mathcal{D}_{\text{dist}}$}
    \State $\hat{y}_{\text{teacher}} \gets \mathcal{M}_{\text{teacher}}(\mathbf{x})$
    \State $\hat{y}_{\text{student}} \gets \mathcal{M}_{\text{student}}(\mathbf{x})$
    \State $\mathcal{L}_{\text{KL}} \gets 
        \text{KL}(\hat{y}_{\text{student}} \| \hat{y}_{\text{teacher}})$ 
        \Comment{Output logit distillation}
    \State $\mathcal{L}_{\text{feat}} \gets \sum_{\ell} 
        \text{MSE}(f_{\ell}^{\text{student}}, f_{\ell}^{\text{teacher}})$ 
        \Comment{Per-layer feature alignment}
    \State $\mathcal{L} \gets \lambda_{\text{KL}} \mathcal{L}_{\text{KL}} + 
        \lambda_{\text{feat}} \mathcal{L}_{\text{feat}}$
    \State Update $\{B_{\text{student}}'\}, \{A_{\text{student}}'^{rs}\}$ 
        via gradient descent on $\mathcal{L}$
\EndFor
\State \Return $\mathcal{M}_{\text{student}}$
\end{algorithmic}
\end{algorithm}

\subsection{LLaMA Compression}
\label{sec:llama_compression_alg}

For LLaMA models, compression operates differently from the ViT setting. \cref{alg:crisp_compression} first performs data-free basis reduction by computing importance scores from template norms and coefficient sparsity, clustering basis vectors via importance-weighted $k$-means, and merging clusters with variance-aware 
rescaling. A short calibration stage (Stage 2) refines the compressed model using a weighted combination of weight reconstruction and language modeling loss, yielding further gains (\cref{tab:compression_llama_peft}).

\begin{algorithm}
\caption{CRISP-Llama Compression Pipeline}
\label{alg:crisp_compression}
\begin{algorithmic}[1]
\Require CRISP model with template banks $\{B_g^{(p)}\}$, coefficients $\{A_i^{(p)}\}$, per-projection compression rates $\{\rho_p\}$, calibration data $\mathcal{D}_{\text{cal}}$ (optional)
\Ensure Compressed model with reduced ranks $r'_p = \lfloor r(1-\rho_p) \rfloor$

\State \textbf{Stage 1: Data-Free Basis Reduction}
\For{each group $g$, projection $p \in \{\text{up, gate, down, q, k, v, o}\}$}
    \State $w \gets \textsc{Importance}(B_g^{(p)}, \{A_i^{(p)}\}, \mathcal{D}_{\text{cal}})$ \Comment{Eq. 1 + optional variance}
    \State $\pi \gets \textsc{Cluster}(B_g^{(p)}, \{A_i^{(p)}\}, w, r'_p)$ \Comment{KMeans + random projection}
    \State $B_g^{(p)} \gets \textsc{Merge}(B_g^{(p)}, \pi, w)$ \Comment{Weighted avg. + rescaling}
    \State $\{A_i^{(p)}\} \gets \textsc{Aggregate}(\{A_i^{(p)}\}, \pi)$ \Comment{Sum per cluster}
\EndFor

\State \textbf{Stage 1b: Coefficient Re-solve} \Comment{Least-squares: $A_i^{(p)} \gets W_{\text{teacher}}^{(p)} (B_g^{(p)})^\dagger$}

\State \textbf{Stage 2: Activation Calibration}
    \For{$e = 1$ to $E_{\text{calib}}$}
        \State $\mathcal{L} \gets w_1 \sum_{i,p} \|W_i^{(p)} - W_{\text{teacher}}^{(p)}\|_F^2 + w_2 \mathcal{L}_{\text{LM}}$
        \State Update $\{B_g^{(p)}\}, \{A_i^{(p)}\}$ via gradient descent
    \EndFor

\State \Return Compressed model
\end{algorithmic}

\textbf{Sub-procedures:}
\begin{algorithmic}[1]
\Function{Importance}{$B, \{A_i\}, \mathcal{D}$}
    \State $w \gets \|B\|_2 + \lambda \sum_i \|A_i\|_1$ \Comment{Per Eq. 1; boost by $\sigma^2$ if $\mathcal{D}$ provided}
    \State \Return $w$
\EndFunction

\Function{Cluster}{$B, \{A_i\}, w, r'$}
    \State $\Phi \gets [\text{RandomProj}(B); \text{concat}(A_i)]$ weighted by $w$
    \State \Return KMeans$(\Phi, r')$
\EndFunction

\Function{Merge}{$B, \pi, w$}
    \State \Return Importance-weighted average with variance rescaling
\EndFunction

\Function{Aggregate}{$\{A_i\}, \pi$}
    \State \Return Sum coefficients per cluster
\EndFunction
\end{algorithmic}
\end{algorithm}

\subsection{Joint MC+PEFT Pipeline}
\cref{alg:joint} presents the complete pipeline for simultaneous 
compression and task adaptation. Compression (ViT: Alg.~\ref{alg:compression}; 
LLaMA: Alg.~\ref{alg:crisp_compression}) is applied first, after which 
basis matrices are frozen and only mixer matrices are updated for 
downstream tasks via Alg.~\ref{alg:peft}. This unified pipeline achieves 
state-of-the-art on both MC (Tab.~2, main paper; \cref{tab:compression_llama_peft}) 
and PEFT (Tab.~1, main paper; \cref{tab:llama_results}) without requiring 
separate optimization procedures.
\begin{algorithm}
\caption{CRISP Joint MC+PEFT}
\label{alg:joint}
\begin{algorithmic}[1]
\Require Pretrained weights $\{W_p^i\}$, target compression rate, downstream task $\mathcal{D}$
\Ensure Compressed and task-adapted model
\State $\{B'\}, \{A'^{rs}\} \gets \text{NeuralMimicry}(\{W_p^i\}, r_{\text{full}}, s_{\text{full}})$ \Comment{Alg.~\ref{alg:neural_mimicry}}
\State $\mathcal{M}_{\text{student}} \gets \text{Compress}(\mathcal{M}_{\text{teacher}}, r_{\text{target}}, s_{\text{target}})$ \Comment{Alg.~\ref{alg:compression}}
\State Freeze $\{B_{\text{student}}'\}$
\State $\{A_{\text{adapted}}'^{rs}\} \gets \text{PEFT}(\mathcal{D}, \{B_{\text{student}}'\}, \{A_{\text{student}}'^{rs}\})$ \Comment{Alg.~\ref{alg:peft}}
\State \Return $\mathcal{M}_{\text{student}}$ with $\{B_{\text{student}}'\}, \{A_{\text{adapted}}'^{rs}\}$
\end{algorithmic}
\end{algorithm}
\section{Datasets}

\textbf{PEFT Evaluation (ViT).} \cref{tab:vtab1k} details the 19 
tasks in the VTAB-1K~\cite{zhai2019large} benchmark used to evaluate PEFT 
on ViT-S/16. Tasks are grouped into three categories: \textcolor{blue!70}{Natural} 
(7 tasks), \textcolor{green!70}{Specialized} (4 tasks, including medical and 
satellite imagery), and \textcolor{orange!70}{Structured} (8 tasks requiring 
geometric and relational reasoning). Each task provides 1K training samples, 
making it well-suited for evaluating sample efficiency and cross-domain 
generalization.
\smallskip

\begin{table}[t]
\centering
\caption{Details of VTAB-1K~\cite{zhai2019large} Benchmark used for PEFT 
on ViT-S/16~\cite{DBLP:journals/corr/abs-2010-11929}.}
\setlength{\tabcolsep}{3pt}
\label{tab:vtab1k}
\begin{tabular}{l c c c c}
\hline
{Dataset} & {\#Cat} & {\#Train} & {\#Val} & {\#Test} \\
\hline
\rowcolor{blue!15}   CIFAR100    & 100 & 800/1000 & 200 & 10000 \\
\rowcolor{blue!15}   Caltech101  & 102 & 6084     & --  & --    \\
\rowcolor{blue!15}   DTD         & 47  & 1880     & --  & --    \\
\rowcolor{blue!15}   Flower102   & 102 & 6149     & --  & --    \\
\rowcolor{blue!15}   Pets        & 37  & 3669     & --  & --    \\
\rowcolor{blue!15}   SVHN        & 10  & 26032    & --  & --    \\
\rowcolor{blue!15}   Sun397      & 397 & 21750    & --  & --    \\
\hline
\rowcolor{green!15}  Camelyon    & 2   & 800/1000 & 200 & 32768 \\
\rowcolor{green!15}  EuroSAT     & 10  & 5400     & --  & --    \\
\rowcolor{green!15}  Resisc45    & 45  & 6300     & --  & --    \\
\rowcolor{green!15}  Retinopathy & 5   & 42670    & --  & --    \\
\hline
\rowcolor{orange!15} Clevr-Count & 8   & 800/1000 & 200 & 15000 \\
\rowcolor{orange!15} Clevr-Dist  & 6   & 15000    & --  & --    \\
\rowcolor{orange!15} DMLab       & 6   & 22735    & --  & --    \\
\rowcolor{orange!15} KITTI-Dist  & 4   & 711      & --  & --    \\
\rowcolor{orange!15} dSpr-Loc    & 16  & 73728    & --  & --    \\
\rowcolor{orange!15} dSpr-Ori    & 16  & 73728    & --  & --    \\
\rowcolor{orange!15} sNORB-Azim  & 18  & 12150    & --  & --    \\
\rowcolor{orange!15} sNORB-Ele   & 9   & 12150    & --  & --    \\
\hline
\end{tabular}
\end{table}

\noindent\textbf{Compression Evaluation (ViT).} \cref{tab:compdataset} details the six fine-grained classification benchmarks used for ViT-B/16 compression evaluation, corresponding to the results in Tab.~2 of the main paper and \cref{tab: taskwise,tab: compression_ablate}. These datasets span a wider range of training budgets than VTAB-1K, allowing us to assess compression quality under less constrained finetuning conditions.

\begin{table}[!htbp]
\centering
\caption{{Details on the Benchmarking Datasets used for fine-tuning Compressed ViT-B/16~\cite{DBLP:journals/corr/abs-2010-11929}  }}
\label{tab:compdataset}
\begin{tabular}{lll}
\toprule
Dataset                                & Classes & \#Sample \\ \midrule
Oxford Flowers~\citep{Nilsback08} & 102              & 6553              \\ 
FGVC Aircrafts~\citep{maji13fine-grained} & 55              & 10001             \\
MIT Scenes~\citep{5206537}     & 67               & 15614             \\ 

CIFAR100 ~\citep{Krizhevsky09}  & 100              & 60000              \\ 
CIFAR10 ~\citep{Krizhevsky09}  & 10             & 60000              \\ 
CUBs (Birds)~\citep{WahCUB_200_2011}          & 200              & 11789             \\ 
 \bottomrule
\end{tabular}
\end{table}

\section{Ablation Studies}
\label{sec:ablation_supp}

We conduct ablation studies to validate key design choices in CRISP's architecture, focusing on three critical components: constraint placement, initialization strategies, and reconstruction loss functions during neural mimicry.
\smallskip

\begin{figure}[t]
    \centering
    \includegraphics[width=\linewidth]{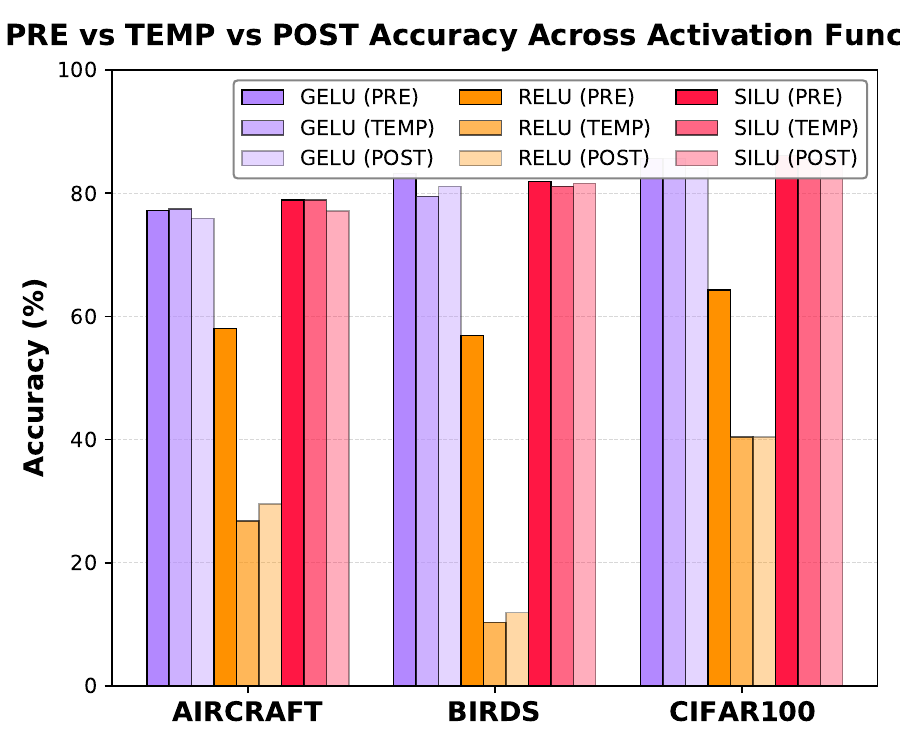}
    \vspace{-4mm}
    \caption{Impact of regularization constraint placement across PRE, POST, and TEMP configurations. PRE (our method) achieves the most consistent performance, while ReLU causes severe degradation due to weight sparsification.}
    \label{fig:activations}
\end{figure}

\noindent\textbf{Regularization via Coefficient Constraints.} 
\cref{fig:activations} compares three strategies for where to apply the regularization constraint in the CRISP transformation: PRE applies the constraint to the mixer before combining $\sigma(A'^{rs}) \odot A'^{rs}$), POST applies it after reconstruction ($\phi(B' A'^{rs})$), and TEMP applies it to the basis matrices ($B' \phi(A'^{rs})$). Our formulation (PRE with SiLU-style gating) consistently outperforms alternatives across all benchmarks, providing implicit regularization on the mixer coefficients without constraining the final weight space. POST and TEMP configurations can slightly hurt performance as they constrain the output weight space directly, which acts as a hard constraint on the layer weights as 
discussed in Sec.~3.1 of the main paper. Notably, ReLU performs catastrophically across all placements due to excessive sparsification of the mixer matrices, consistent with the ablation results in Tab.~4 of the main paper. GELU shows competitive performance to SiLU, but SiLU's smooth gating better balances expressivity and regularization without introducing additional hyperparameters.
\smallskip

\begin{figure}[t]
    \centering
    \includegraphics[width=\linewidth]{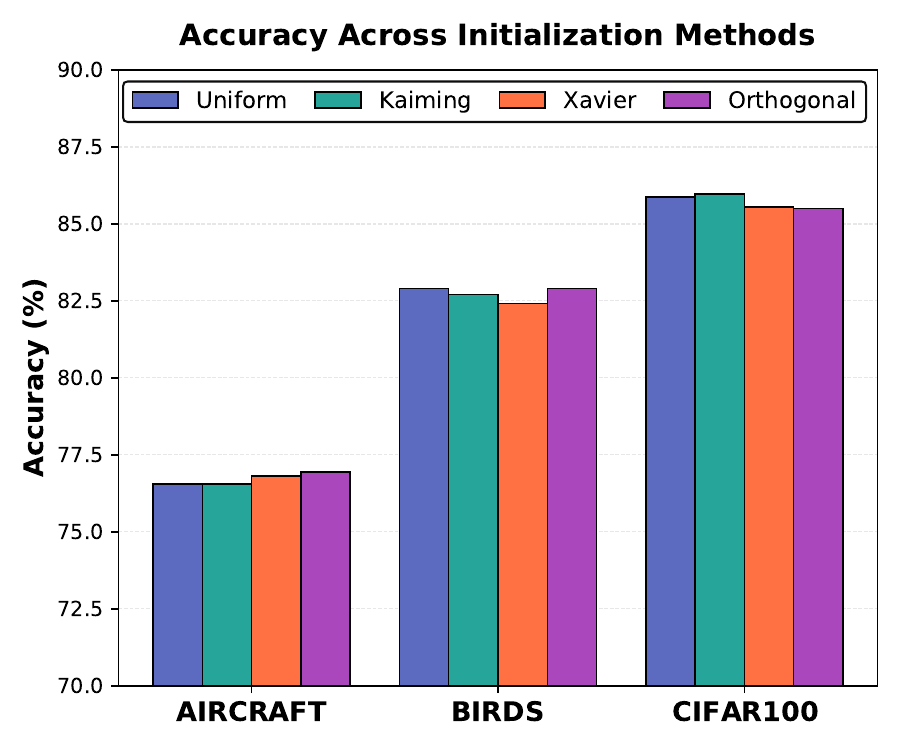}
    \vspace{-4mm}
    \caption{\textbf{Robustness to initialization methods.} We evaluate four standard initialization schemes (Uniform, Kaiming, Xavier, Orthogonal) for the mixer matrices $A'^{rs}$ during both neural mimicry retrofitting and subsequent task adaptation.}

    \label{fig:init}
\end{figure}

\noindent\textbf{Initialization methods.} \cref{fig:init} evaluates four initialization strategies for the mixer matrices $A'^{rs}$ during \cref{alg:neural_mimicry}: uniform, Kaiming, Xavier, and orthogonal. Results show remarkable robustness across initialization schemes, with all methods achieving within 1\% of each other on most tasks. This insensitivity to initialization validates that the neural mimicry objective effectively guides the learning process regardless of starting point. The slight advantage of Kaiming and orthogonal initializations on certain tasks motivated our choice of orthogonal initialization as mentioned in the paper, but practitioners can confidently use simpler schemes without significant performance degradation.
\smallskip

\begin{figure}[t]
    \centering
    \includegraphics[width=\linewidth]{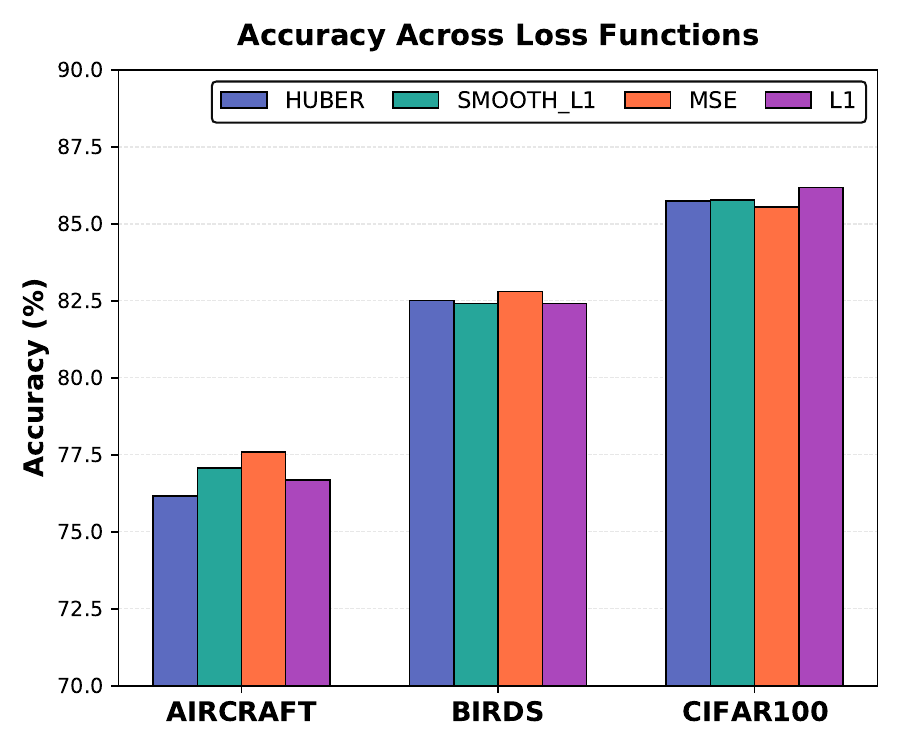}
    \vspace{-4mm}
    \caption{\textbf{Effect of reconstruction loss functions during neural mimicry.} We compare four loss functions (Huber, Smooth-L1, MSE, L1) used in the neural mimicry stage (Equation~5 of main paper) for retrofitting pretrained weights into CRISP's basis-mixer decomposition. }

    \label{fig:loss}
\end{figure}

\noindent\textbf{Loss functions for neural mimicry.} \cref{fig:loss} compares reconstruction losses in \cref{alg:neural_mimicry}: Huber, smooth-L1, MSE, and L1. All robust losses (Huber, smooth-L1) perform comparably, with smooth-L1 showing marginal advantages on fine-grained tasks like Aircraft. The consistent performance across loss functions suggests that the choice of reconstruction objective is less critical than the overall factorization framework, though smooth-L1's robustness to outliers during weight decomposition motivated its adoption in our implementation. Notably, L1 loss shows competitive or superior performance on some tasks, indicating potential for further exploration of sparsity-inducing objectives during retrofitting.
\smallskip


\noindent\textbf{Compression ablations.}
 \cref{tab: compression_ablate} evaluates design choices for compressing ViT-B/16 by 50\% across six benchmarks (see \cref{tab:compdataset}). Neural mimicry alone (Eq.\ 5 on main paper) significantly underperforms using distillation, demonstrating that weight-space reconstruction without data is insufficient for aggressive compression. In contrast, distillation boosts accuracy by 31\%, validating our two-stage approach. We also show that using SVD initialization is important (replacing it with random orthogonal initialization) as it provides a boost of almost 4\%, confirming that eigenvector-based warm-starting provides a stronger initialization for the compressed parameter space.
\smallskip
 
 \begin{table*}[t]
\centering
\caption{Compression results on ViT-B/16~\cite{DBLP:journals/corr/abs-2010-11929} with 50\% weight compression across six diverse tasks (See Table~\ref{tab:compdataset}).  We find that using the distillation loss with SVD initialization described in Sec.\ 3.2 of our paper provides best performance.}

\label{tab: compression_ablate}
\begin{tabular}{lcccccccc}
\hline
Compression & Params(\%) & Flowers & Aircraft & Scene & CFR100 & CFR10 & Birds & Avg \\
\hline
Neural Mimicry (Eq.\ 5 of main paper)  & 44M & 83.9 & 56.3 & 51.2 & 49.8  & 79.4 & 26.3 & 57.8 \\
Distillation Loss & 44M & {99.0} & {89.5} & 81.8 & {86.2} & 97.4 & {79.1} & \textbf{88.8} \\
\hspace{2mm} w/o SVD Init &  44M & 98.8 & 85.9 & 77.4 & 82.0 & 95.3 & 71.6 & 85.1 \\
\bottomrule
\end{tabular}
\end{table*}

\noindent\textbf{Additional ViT Compression Results}
\cref{tab: taskwise} reports performance on 75\% compression to supplement our 50\% compression results from our main paper.  We find that \method{} outperforms other PR methods like RECAST~\cite{tasnimRECAST2025} by 10\% on MC.  This gap is increased when combined with PEFT methods, where we boost performance by 11\%.  This helps further highlight the benefits of our approach over prior general PR approaches.
\smallskip
\begin{table*}[t]
\centering
\caption{Compression results on ViT-B/16~\cite{DBLP:journals/corr/abs-2010-11929}  at 75\% parameter reduction evaluated on six fine-grained classification benchmarks (see Tab.\ 2 of the main paper for 50\% reduction). \textbf{Upper section:} post-compression accuracy with only classifier adaptation. \textbf{Lower section:} MC+PEFT combinations demonstrate compressed models as initialization for downstream tasks. 
We find that \method{} outperforms prior work by up to 11\%.}
\label{tab: taskwise}
\begin{tabular}{lccccccccc}
\hline
Compression & PEFT & Params(\%) & Flowers & Aircraft & Scene & CFR100 & CFR10 & Birds & Avg \\
\hline
\rowcolor{gray!20}ViT-B/16~\cite{DBLP:journals/corr/abs-2010-11929} & -- & 86M & 96.7 & 70.9 & 84.5 & 76.3 & 97.0 & 84.6 & 85.0 \\
\midrule
\rowcolor{red!15}SVD  & -- &21M & 70.7 & 25.7 & 33.2 & 39.6  & 62.5 & 8.6 & 40.0 \\
\rowcolor{red!15}RECAST~\cite{tasnimRECAST2025} & -- & 21M & 90.0 & 59.1 & 67.2 & 67.1  & 85.3 & 52.9 & 70.2 \\
\rowcolor{red!40}\method{} (ours) & --  & 21M & 95.1 & 73.0 & 77.7 & 74.1  & 89.7 & 73.6 & \textbf{80.5} \\
\midrule 
\rowcolor{blue!15}SVD &  Eigenvalues  & 21M & 87.8 & 55.9 & 62.1 & 65.0  & 85.3 & 40.1 & 66.0 \\
\rowcolor{blue!15}RECAST~\cite{tasnimRECAST2025} & RECAST & 21M & 94.7 & 67.4 & 71.0 & 71.4  & 89.5 & 58.9 & 75.5 \\
\rowcolor{blue!40}\method{} (ours) & \method{} (ours)& 21M & 98.8 & 85.2 & 79.6 & 82.6  & 95.9 & 75.1 & \textbf{86.2} \\
\bottomrule
\end{tabular}
\end{table*}

\section{Experiments on Large Language Models}
\label{sec:llama_discussion}

Following standard protocols, we initialize LLaMA models with our \method{} reparameterization via neural mimicry and then fine-tune only the mixer matrices while keeping basis matrices frozen. We evaluate at two parameter budgets: an ultra-low regime with approximately 0.004\% trainable parameters and a moderate regime at 0.01\%, both substantially lower than conventional PEFT methods which typically use around 0.7-0.8\% of base model parameters.

\cref{tab:llama_results} demonstrates that \method{} is on par or better than its competitors with an order of magnitude fewer parameters.  We also show that we can reduce the number of trainable parameters by another order of magnitude with only a minimal impact to performance.  Note that prior work has shown PR methods like \method{} can also be composed with methods like HiRA and DoRA for further gains~\cite{tasnimRECAST2025}. Further, most of these methods we compare to can only be applied to PEFT, whereas our approach can be used for compression as well, easing the implementation costs.

\begin{table*}[t]
\centering
\small
\caption{Comparison of PEFT methods on commonsense reasoning benchmarks. Results from LoRA and DoRA are taken from~\citet{10.5555/3692070.3693369}, HiRA results are from~\citet{huang2025hira}.  We find that \method{} is on par or better than custom PEFT methods while using an order of magnitude fewer parameters.  Further, \method{} can also support MC, as we show in~\cref{tab:compression_llama_peft}, demonstrating its ability to generalize to more PR tasks than prior PEFT methods.}
\label{tab:llama_results}. 
\begin{tabular}{l@{\hspace{3pt}}l@{\hspace{3pt}}c@{\hspace{4pt}}c@{\hspace{4pt}}c@{\hspace{4pt}}c@{\hspace{4pt}}c@{\hspace{4pt}}c@{\hspace{4pt}}c@{\hspace{4pt}}c@{\hspace{4pt}}c@{\hspace{4pt}}c}
\toprule
\multirow{2}{*}{{Model}} & \multirow{2}{*}{{PEFT}} & \multirow{2}{*}{{Params(\%)}} & \multicolumn{8}{c}{{Accuracy ($\uparrow$)}} \\
\cmidrule(lr){4-11}
& & & {BQ} & {PIQ} & {SIQ} & {Hell.} & {Win.} & {ARC-e} & {ARC-c} & {OBQ} & {Avg.} \\
\midrule
ChatGPT & --- & --- & 73.1 & 85.4 & 68.5 & 78.5 & 66.1 & 89.8 & 79.9 & 74.8 & 77.0 \\
\midrule
& LoRA~\cite{hu2022lora} & 0.83 & 69.8 & 79.9 & 79.5 & 83.6 & 82.6 & 79.8 & 64.7 & 81.0 & 77.6 \\
Llama2-7B & DoRA$_{\text{half}}$~\cite{10.5555/3692070.3693369} & 0.42 & 72.0 & 83.1 & 79.9 & 89.1 & 83.0 & 84.5 & 71.0 & 81.2 & 80.5 \\
& DoRA~\cite{10.5555/3692070.3693369} & 0.84 & 71.8 & 83.7 & 76.0 & 89.1 & 82.6 & 83.7 & 68.2 & 82.4 & 79.7 \\
& HiRA~\cite{huang2025hira} & 0.83 & 71.2 & 83.4 & 79.5 & 88.1 & 84.0 & 86.7 & 73.8 & 84.6 & \textbf{81.4} \\
\cmidrule(lr){2-11}
& \method{} & $0.004$ & 68.9 & 81.4 & 80.4 & 91.0 & 80.1 & 84.1 & 69.1 & 73.6 & 78.6\\
& \method{} & $0.01$ & 69.5 & 81.8 & 80.4 & 91.7 & 83.6 & 84.6 & 69.2 & 78.2 & 80.0 \\

\midrule

& LoRA~\cite{hu2022lora} & 0.70 & 70.8 & 85.2 & 79.9 & 91.7 & 84.3 & 84.2 & 71.2 & 79.0 & 80.8 \\
Llama3-8B & DoRA$_{\text{half}}$~\cite{10.5555/3692070.3693369} & 0.36 & 74.5 & 88.8 & 80.3 & 95.5 & 84.7 & 90.1 & 79.1 & 87.2 & 85.0 \\
& DoRA~\cite{10.5555/3692070.3693369} & 0.71 & 74.6 & 89.3 & 79.9 & 95.5 & 85.6 & 90.5 & 80.4 & 85.8 & 85.2 \\
 &HiRA~\cite{huang2025hira} & 0.70 & 75.4 & 89.7 & 81.2 & 95.4 & 87.7 & 93.3 & 82.9 & 88.3 & \textbf{86.7} \\
\cmidrule(lr){2-11}
& \method{} & $0.004$ & 72.3 & 87.6 & 81.5 & 94.3 & 87.0 & 91.5 & 79.1 & 83.8 & 84.7 \\
& \method{} & $0.01$ & 73.6 & 89.1 & 80.8 & 94.8 & 85.7 & 93.1 & 83.0 & 87.6 & 86.0 \\
\bottomrule
\end{tabular}
\end{table*}
\cref{alg:crisp_compression} describes the LLaMA compression pipeline, 
which reduces basis rank via importance-weighted clustering and a short 
calibration stage using language modeling loss. \cref{tab:compression_llama_peft} 
evaluates this against compression+PEFT combinations on commonsense reasoning 
at 30\% parameter reduction. We follow the same procedure outlined in the main paper for adapter-finetuning and evaluation.

In the compression-only setting (main paper Tab.~3), \method{} outperforms all baselines by around 5 points - a gap consistent with the ViT results in Tab.~2 of the main paper. In the \cref{tab:compression_llama_peft}, we evaluate compressed models as initializations for downstream PEFT. \method{} with coefficient tuning achieves surpasses Basis-Sharing~\cite{wang2025basis} and DFJR~\cite{DBLP:journals/corr/abs-2402-16319} combinations by a significant margin. PruneNet~\cite{sengupta2025you} combined with LoRA~\cite{hu2022lora} or DoRA~\cite{10.5555/3692070.3693369} achieves higher accuracy, though these combinations use substantially more trainable parameters during adaptation than \method{}'s lightweight coefficient tuning. This highlights a parameter-accuracy tradeoff: \method{} provides a unified compress-once-adapt-freely framework at a lower adaptation cost, whereas pruning-based methods require a separate PEFT pipeline with a larger parameter budget to reach their best performance.  That said, prior work has shown methods like \method{} can be combined with alternatives like LoRA for improved PEFT performance~\cite{tasnimRECAST2025}, which we leave to future work.


\begin{table*}[t]
\centering
\caption{LLaMA3.2-1B~\cite{dubey2024llama3} MC+PEFT results at 30\% parameter reduction across seven commonsense reasoning benchmarks. All baselines pair a compression method with a separate PEFT method, whereas \method{} compresses and adapts within the same factorized framework using lightweight coefficient tuning.}
\label{tab:compression_llama_peft}
\begin{tabular}{l@{\hspace{4pt}}l@{\hspace{6pt}}l@{\hspace{6pt}}c@{\hspace{6pt}}c@{\hspace{6pt}}c@{\hspace{6pt}}c@{\hspace{6pt}}c@{\hspace{6pt}}c@{\hspace{6pt}}c@{\hspace{6pt}}c}
\toprule
& \textbf{Compression} & \textbf{PEFT} & \multicolumn{8}{c}{Accuracy ($\uparrow$)} \\
\cmidrule(lr){4-11}
& & & BQ & PIQ & Hell. & Wino & ARC-e & ARC-c & OBQ & Avg. \\
\midrule
\rowcolor{gray!20} 
& LLaMA3.2-1B~\cite{dubey2024llama3} & -- & 60.0 & 90.0 & 30.0 & 70.0 & 70.0 & 30.0 & 10.0 & 51.4 \\
\midrule

\rowcolor{green!15} 
& Basis-Sharing~\cite{wang2025basis} & LoRA~\cite{hu2022lora} & 37.8 & 52.8 & 26.8 & 48.6 & 28.2 & 19.3 & 16.2 & 32.8 \\

\rowcolor{green!15} 
& Basis-Sharing~\cite{wang2025basis} & DoRA~\cite{10.5555/3692070.3693369} & 37.8 & 53.8 & 26.8 & 51.0 & 28.9 & 19.2 & 13.2 & 33.0 \\

\rowcolor{green!15} 
& DFJR~\cite{DBLP:journals/corr/abs-2402-16319} & LoRA~\cite{hu2022lora} & 48.9 & 53.5 & 26.4 & 50.3 & 28.6 & 20.6 & 14.4 & 34.7 \\

\rowcolor{green!15} 
& DFJR~\cite{DBLP:journals/corr/abs-2402-16319} & DoRA~\cite{10.5555/3692070.3693369} & 57.2 & 54.7 & 27.3 & 50.9 & 29.2 & 19.8 & 13.6 & 36.1 \\
\rowcolor{green!15} 
& PruneNet~\cite{sengupta2025you} & LoRA~\cite{hu2022lora}& 62.4 & 57.8 & 32.3 & 54.6 & 39.4 & 22.1 & 16.8 & 40.8  \\

\rowcolor{green!15} 
& PruneNet~\cite{sengupta2025you} & DoRA~\cite{10.5555/3692070.3693369} & 62.3 & 60.3 & 32.6 & 55.4 & 40.4 & 21.5 & 17.6 & 41.4 \\

\rowcolor{green!40} 
& \method{} (ours) & \method{} (ours) & 57.4 & 53.4 & 27.9 & 51.4 & 30.8 & 23.4 & 26.6 & 38.7 \\

\bottomrule
\end{tabular}
\end{table*}

{
    \small
    \bibliographystyle{ieeenat_fullname}
    \bibliography{main}
}


\end{document}